%% file: main.tex
\documentclass[10pt,twocolumn,letterpaper]{article}

\usepackage{cvpr}              

\definecolor{cvprblue}{rgb}{0.21,0.49,0.74}
\usepackage[pagebackref,breaklinks,colorlinks,allcolors=cvprblue]{hyperref}
\usepackage{amsmath}
\usepackage{xcolor}
\usepackage{multirow}
\usepackage{booktabs}
\usepackage{tabularx} 
\usepackage{graphicx}
\usepackage[export]{adjustbox} 
\usepackage{graphicx} 
\usepackage{caption}
\usepackage{adjustbox}
\usepackage{fix-cm}
\usepackage[accsupp]{axessibility}
\usepackage[table]{xcolor}
\usepackage{amssymb}
\usepackage{pifont}
\usepackage[accsupp]{axessibility}

\usepackage{tikz}
\usetikzlibrary{arrows,positioning}

\def\paperID{} % *** Enter the Paper ID here
\def\confName{CVPR}
\def\confYear{2026}

\title{ConsisVLA-4D: Advancing Spatiotemporal Consistency in Efficient 3D-Perception and 4D-Reasoning for Robotic Manipulation}

\author{
    Wei Li$^1$, ~~Jizhihui Liu$^1$, ~~Li Yixing$^1$, ~~Junwen Tong$^2$, ~~Rui Shao$^{1, 3}$\thanks{Corresponding author.}, ~~Liqiang Nie$^1$\\
    $^1$Harbin Institute of Technology, Shenzhen \\
    $^2$State Key Laboratory of Mobile Network and Mobile Multimedia Technology, ZTE, China\\
    $^3$Shenzhen Loop Area Institute\\
    {\tt liwei2024@stu.hit.edu.cn \quad shaorui@hit.edu.cn}\\
\texttt{
{\url{https://github.com/JiuTian-VL/ConsisVLA-4D}}}
}

\begin{document}

\maketitle

% \twocolumn[{
% \maketitle

% \begin{center}
% \includegraphics[width=\linewidth]{fig/freecompress-intro.png}
% \end{center}
% \vspace{-0.5cm}
% \captionsetup{type=figure}
% \captionof{figure}{%
% VGGT is a large feed-forward transformer with minimal 3D-inductive biases trained on a trove of 3D-annotated data.
% It accepts up to hundreds of images and predicts cameras, point maps, depth maps, and point tracks for all images at once in less than a second, which often outperforms optimization-based alternatives without further processing. 
% }\label{fig:intro}
% \vspace{0.3cm}
% }]

\input{sec/0_abstract}    
\input{sec/1_intro}

\input{sec/2_related}

\input{sec/3_pre}
\input{sec/4_consis}

\input{sec/5_exp}

\input{sec/6_conclusion}
\input{sec/X_suppl}

% \newpage

{
    \small
    \bibliographystyle{ieeenat_fullname}
    \bibliography{main}
}


\end{document}

%% file: sec/0_abstract.tex
\begin{abstract}
Current Vision-Language-Action (VLA) models primarily focus on mapping 2D observations to actions but exhibit notable limitations in spatiotemporal perception and reasoning: 
1) spatial representations often rely on additional sensors, introducing substantial computational overhead;
2) visual reasoning is typically limited to future-frame prediction, lacking alignment with the instruction-grounded scene and thus compromising spatiotemporal consistency.
To address these challenges, we propose \textbf{ConsisVLA-4D}, a unified and efficient framework that enhances spatiotemporal consistency in 3D-Perception and 4D-Reasoning. Specifically, we design:
\textbf{1) CV-Aligner}, which ensures \textbf{C}ross-\textbf{V}iew object semantic consistency via filtering instruction-relevant regions and aligning object identities across multiple viewpoints;
\textbf{2) CO-Fuser}, which guarantees \textbf{C}ross-\textbf{O}bject spatial geometric consistency by eliminating spatial relation ambiguities between objects across views using compact latent representations.
Building upon these, we introduce 
\textbf{3) CS-Thinker} to achieve \textbf{C}ross-\textbf{S}cene spatiotemporal consistency as actions unfold. It learns implicit knowledge of local dynamics from object-semantic tokens of CV-Aligner and global depth from geometric tokens of CO-Fuser, thereby enhancing efficient visual reasoning under scene variations.
Extensive experiments demonstrate that, benefiting from its efficient spatiotemporal consistency design, ConsisVLA-4D achieves \textbf{21.6\%} and \textbf{41.5\%} performance improvements, along with \textbf{2.3$\times$} and \textbf{2.4$\times$} inference speedups compared to OpenVLA on the LIBERO benchmark and real-world platforms, respectively.
\end{abstract}

%% file: sec/1_intro.tex
\begin{figure}[t]
  \centering
   \includegraphics[width=1.0\linewidth]{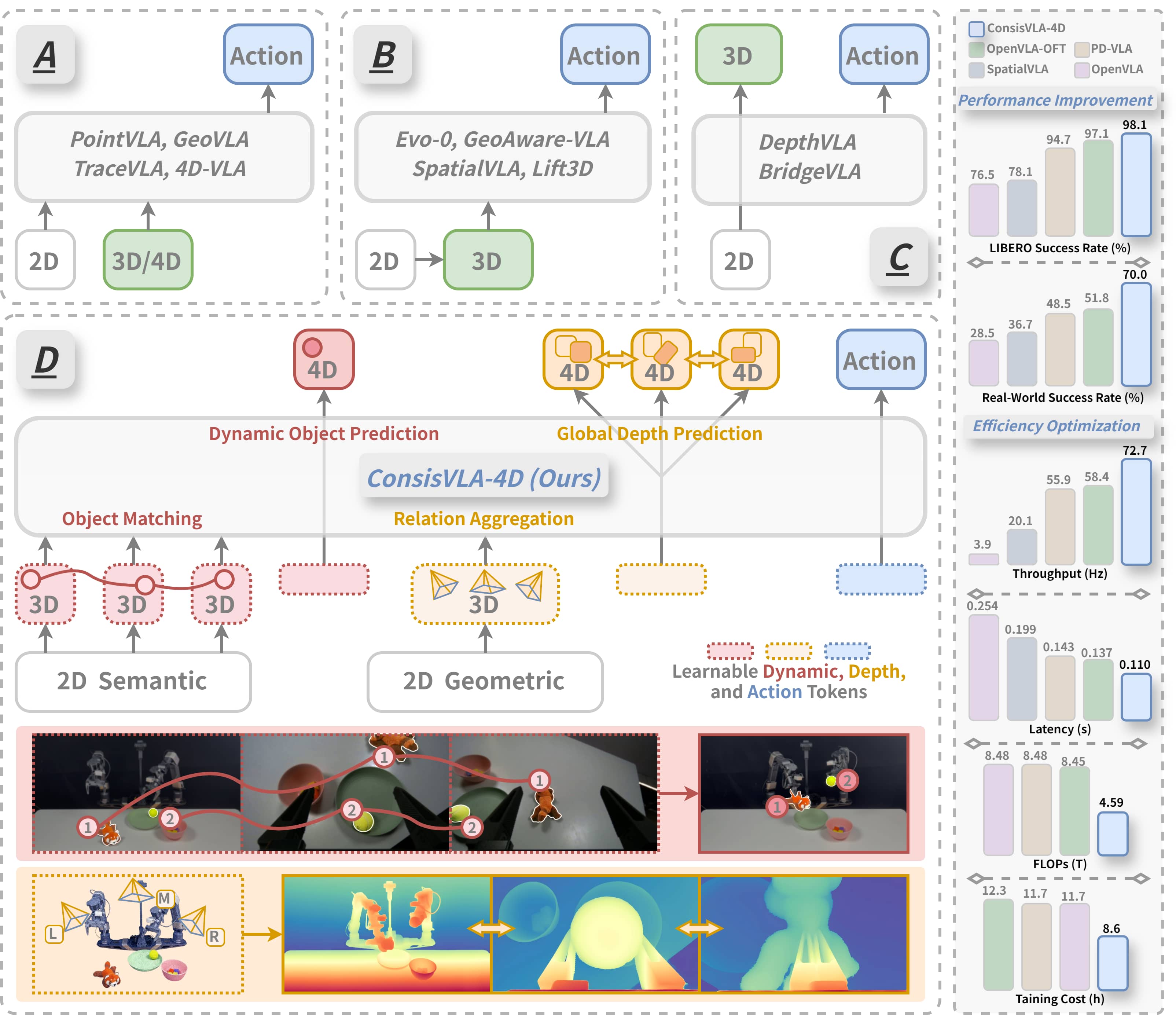}
   % \vspace{-1.7em}
    \caption{\textbf{Comparison with Existing Paradigms.}
    Beyond conventional 2D visual inputs, \textbf{\underline{Para.~A}} employs explicit 3D/4D inputs (e.g., point clouds, depth maps, historical frames), \textbf{\underline{Para.~B}} projects 2D inputs into 3D space, and \textbf{\underline{Para.~C}} predicts 3D representations from 2D observations. In contrast, we extend the paradigm from 3D-Perception to 4D-Reasoning within a unified framework (\textbf{\underline{Para.~D}}):
   \textbf{1)} CV-Aligner extracts instruction-related and cross-correlated spatial objects; 
   \textbf{2)} CO-Fuser aggregates multi-view geometric relation; 
   \textbf{3)} CS-Thinker infers actions based on implicit knowledge of future dynamic objects and global depth. 
   \textbf{ConsisVLA-4D} achieves spatiotemporal consistency using only about \textbf{1/8} of the original visual inputs, enabling robust and efficient robotic manipulation, as illustrated on the right.}
   % \vspace{-10pt}
   \label{fig:intro}
\end{figure}

% \vspace{-10pt}
\section{Introduction}
\label{sec:intro}
% \vspace{-5pt}

% Vision-Language-Action (VLA) models~\cite{ACT, openvla, oft, pi0, bu2025univla, zhong2025dexgraspvla} have made significant progress in enabling robots to perform complex tasks by integrating multimodal information, including visual perception and natural language instructions. Representative works such as RT-2~\cite{Rt-2}, Octo~\cite{octo}, OpenVLA~\cite{openvla} and $\pi$-series~\cite{pi0, pi05, fast, driess2025knowledge} have demonstrated the potential of the VLA paradigm in bridging the gap between 2D perception and spatial manipulation, achieving impressive results across diverse tasks. However, despite these advancements, these VLA models still face significant limitations in spatiotemporal perception and reasoning, hindering their performance in real-world scenarios.
Vision-Language-Action (VLA) models~\cite{openvla, oft, pi0, bu2025univla, li2025cogvla, li2025semanticvla, ni2025swiftvla, zhu2026H-GAR, shao2025large, li2026global} have made significant progress in enabling robots to perform complex tasks by integrating visual perception and language instructions. 
Representative works such as RT-2~\cite{Rt-2}, Octo~\cite{octo}, OpenVLA~\cite{openvla}, and $\pi$-series~\cite{pi0, pi05, fast, driess2025knowledge} highlight the potential of the VLA paradigm in bridging the gap between 2D perception and spatial manipulation, achieving impressive results across diverse benchmarks. 
However, despite these advances, current VLA models remain limited in spatiotemporal perception and reasoning. For instance, when confronted with complex object layouts or action-induced scene changes, they often struggle with these tasks, limiting their real-world effectiveness.

The core challenges lie in two aspects.
First, \textbf{3D spatial understanding} remains a significant bottleneck for VLA models. 
Although some methods attempt basic 2D-to-3D mapping~\cite{lin2025evo, geoaware, spatialvla, ni2025wonderturbo}, they fail to capture the full structural complexity of the 3D world, often leading to object misalignment and spatial relation ambiguity.
Methods~\cite{li2025pointvla, sun2025geovla, zheng2024tracevla} that utilize 3D representations, such as point clouds or depth maps, require additional sensors and incur substantial computational overhead, limiting their applicability in resource-constrained settings.
Second, \textbf{4D spatiotemporal reasoning} remains underdeveloped~\cite{cen2025worldvla, chen2025world4omni, assran2025v, cotvla}. During action execution, spatial scenes evolve dynamically. 
Due to the lack of a comprehensive understanding of current spatial states and insufficient knowledge of evolving scene dynamics, existing methods struggle to build consistent correlations with predicted future scenes. 
% This weakens spatiotemporal consistency, leading to degraded visual reasoning and unstable action generation.
This deficiency undermines spatiotemporal consistency, resulting in impaired visual reasoning and unstable action generation.

% In contrast, humans excel in manipulation tasks due to their efficient integration of spatiotemporal consistency~\cite{blake2011binocular, chen2024integration}: through binocular vision or movement, humans maintain accurate and consistent spatial perception (e.g., object locations and relationships) across different viewpoints. Simultaneously, during manipulation, humans perform visual reasoning (e.g., predicting future states) based on this stable spatial perception, ensuring temporal stability in spatial awareness throughout task execution.
In contrast, humans excel in manipulation tasks owing to their robust and efficient spatiotemporal consistency~\cite{blake2011binocular, chen2024integration}. Through binocular vision or movement, humans maintain accurate and consistent spatial perception (e.g., object locations and relationships) across diverse viewpoints. Simultaneously, humans perform visual reasoning (e.g., predicting future spatial states) based on this stable spatial perception, ensuring temporal stability throughout task execution.

Motivated by human behavior and its relevance to VLA, we aim to address two key questions: 
\textbf{1)} How can we efficiently generate 3D representations from 2D observations without excessive computational overhead? 
\textbf{2)} How can we enhance spatiotemporal consistency to optimize action prediction through 4D visual reasoning?
To this end, we propose \textbf{ConsisVLA-4D}, a unified and efficient framework that enhances spatiotemporal consistency in 3D-perception and 4D-reasoning, as shown in Fig.~\ref{fig:intro}.
Specifically, ConsisVLA-4D consists of three tightly integrated and coherent core components:
\textbf{1) Cross-View Aligner (CV-Aligner)} filters redundant visual representations through Explicit Semantic Object Selection, retaining only instruction-relevant objects. It then aligns object identities across viewpoints using Single-Fusion, producing efficient spatial representations (\textbf{1/8} of the original) that ensure Cross-View object semantic consistency.
\textbf{2) Cross-Object Fuser (CO-Fuser)} processes multiple viewpoints via Group-Fusion to resolve geometric ambiguities in object positions from a single view. It refines spatial relationships between objects through Implicit Geometric Relation Aggregation, yielding compact latent representations (\textbf{1/12-1/8} of the original) that ensure Cross-Object Spatial Geometric Consistency.
\textbf{3) Cross-Scene Thinker (CS-Thinker)} learns two sets of implicit knowledge: dynamic objects in future fixed viewpoints using semantic tokens from CV-Aligner, and global depth across all viewpoints using geometric latent tokens from CO-Fuser. 
During inference, these pre-learned knowledge sets account for less than \textbf{10\%} of the observation-instruction sequence, enabling efficient 4D visual reasoning.
Through this design, ConsisVLA-4D achieves a stable and coherent understanding of dynamic scenes and maintains spatiotemporal consistency throughout action generation.

Our contributions are summarized as follows:

\begin{itemize}
    \item We propose \textbf{ConsisVLA-4D}, an efficient and innovative framework that advances spatiotemporal consistency in 3D-Perception and 4D-Reasoning.
    \item We introduce \textbf{CV-Aligner} and \textbf{CO-Fuser} to ensure cross-view object semantic consistency and cross-object spatial geometric consistency, thereby improving the model’s understanding of complex 3D scenes.
    \item We design \textbf{CS-Thinker} to enhance 4D spatiotemporal reasoning, ensuring the model generates consistent and reliable actions in dynamic environments.
    \item Extensive experiments demonstrate that ConsisVLA-4D, with its advanced spatiotemporal consistency design, significantly outperforms existing methods in both performance and computational efficiency.
\end{itemize}

% \begin{figure*}[t]
%   \centering
%    \includegraphics[width=1.0\linewidth]{fig/freecompress-framework.png}
%    \caption{The whole framework of LION-FS. {Fast Path} enables high-frame-rate video stream reception, allowing real-time determination of whether a response is required. $E_{gen}$ extracts general spatial features from 2 FPS frames, while $E_{ego}$ captures first-person temporal features from 8 FPS frames. These are temporally aligned, weighted through the Token Aggregation Router, and then filtered for redundancy by the Token Dropping Router. {Slow Path} enhances keyframes with 
%    rich information, performing multi-granularity augmentation that includes fine-grained global tokens (Grid Tokens) and action-related local tokens (Box Tokens), which are injected into the Multimodal Thinking Template to guide the assistant in generating more precise responses.}
%    \vspace{-10pt}
%    \label{fig:framework}
% \end{figure*}

%% file: sec/2_related.tex
% \vspace{-5pt}
\section{Related Work}
\label{sec:Related_Work}

% \vspace{-5pt}
% \paragraph{VLM-based VLA Models.}
% Vision-Language Models (VLMs)~\cite{li2023blip, liu2023visual, liu2024improved, llama3, yang2025qwen3, comanici2025gemini} have substantially advanced robotic control and task execution by bridging visual perception and linguistic understanding.  
% RT-2~\cite{Rt-2} introduced the concept of Vision-Language-Action (VLA) modeling based on pre-trained VLMs, and RT-X~\cite{vuong2023open} expanded it to larger scales for improved generalization and reasoning.
% OpenVLA~\cite{openvla}, trained on 970K real robot demonstrations, publicly released a 7B-parameter VLA model, while OpenVLA-OFT~\cite{oft} improved inference efficiency through parallel decoding and action chunking.
% These single-system designs~\cite{cotvla, pdvla, deervla, molevla, hybridvla} inspired a series of follow-up studies. 
% In comparison, dual-system architectures such as GR00T N1~\cite{bjorck2025gr00t}, $\pi_{0}$~\cite{pi0}, $\pi_{0.5}$~\cite{pi05}, and SmolVLA~\cite{shukor2025smolvla} explicitly separate a VLM backbone for high-level reasoning from an action expert for low-level control, enabling more structured end-to-end coordination. 
% However, most of these models remain aligned with the 2D vision-language generation paradigm of VLMs, which limits their ability to perceive spatial information and predict future states that are crucial for accurate and robust robotic manipulation.

% \vspace{-2pt}
\paragraph{VLM-based VLA Models.}
Vision-Language Models (VLMs)~\cite{li2023blip, liu2023visual, liu2024improved, llama3, yang2025qwen3, comanici2025gemini, li2025lion, zhang2025falcon, li2025optimus, li2026optimus3, chen2025less, lyu2026personalalign, cai2025tone, codes2026} and deep learning~\cite{shao2019multi, shao2023detecting, shao2024detecting, zhou2025hiconagent, lyu2025puma, zhu2025emosym, zhu2025uniemo, xie2025gui, shao2017deep, li2025taco, zhang2023tdec, zheng2024deep} have advanced robotic manipulation by bridging visual perception with language.
RT-2~\cite{Rt-2} introduced VLA modeling with pre-trained VLMs, and RT-X~\cite{vuong2023open} scaled it for better generalization.
OpenVLA~\cite{openvla}, trained on 970K robot demonstrations, released a 7B-parameter model, while OpenVLA-OFT~\cite{oft} optimized inference with action chunking.
These single-system models~\cite{cotvla, pdvla, deervla, molevla, hybridvla} sparked numerous subsequent studies. In contrast, dual-system architectures, such as GR00T N1~\cite{bjorck2025gr00t},  CogACT~\cite{ACT}, $\pi_{0}$~\cite{pi0}, and SmolVLA~\cite{shukor2025smolvla}, separate the VLM backbone for high-level reasoning from an action expert for low-level control, fostering structured coordination. 
However, most models remain limited to a 2D vision-language paradigm, restricting spatial awareness and future state prediction needed for robust manipulation.

%-------------------------------------------------------------------------

% \paragraph{Spatiotemporal Modeling Approaches in VLA Models.} Recent works have explored incorporating spatial information into robotic manipulation. 
% Approaches such as LEO~\cite{huang2023embodied}, FP3~\cite{yang2025fp3}, 3D-VLA~\cite{zhen20243d}, PointVLA~\cite{li2025pointvla}, GeoVLA~\cite{sun2025geovla}, Lift3D~\cite{jia2024lift3d}, and 3DS-VLA~\cite{li20253ds} incorporate 3D representations like point clouds and depth maps, while 4D-VLA~\cite{zhang20254d} and TraceVLA~\cite{zheng2024tracevla} further extend inputs to 4D signals such as historical frames and visual traces. 
% However, these methods rely on specialized sensors, which constrain their applicability across diverse robotic platforms, and incur significant computational costs. 
% Alternatively, models such as SpatialVLA~\cite{spatialvla}, BridgeVLA~\cite{li2025bridgevla}, and Spatial Forcing~\cite{spatialforcing2025} attempt to infer 3D spatial structures directly from 2D perception and action prediction. 
% Yet, these 2D-to-3D reasoning paradigms suffer from projection bias, lack of cross-view and cross-object geometric consistency, and vulnerability to occlusion and hierarchical errors.
% Consequently, there is an urgent need for a general and efficient framework for 3D spatial perception modeling that can capture fine-grained spatial relationships and maintain temporal coherence.

% \vspace{-10pt}
\paragraph{Spatiotemporal Modeling Approaches in VLA Models.}
Recent works~\cite{huang2023embodied, yang2025fp3, zhen20243d, li2025pointvla, sun2025geovla, jia2024lift3d, li20253ds, YANG2026114166, YANG2026132599} like 3D-VLA~\cite{zhen20243d}, PointVLA~\cite{li2025pointvla}, GeoVLA~\cite{sun2025geovla}, and Lift3D~\cite{jia2024lift3d} leverage 3D representations, while 4D-VLA~\cite{zhang20254d} and TraceVLA~\cite{zheng2024tracevla} extend this to 4D cues, incorporating historical frames and visual traces.
However, these approaches rely on specialized sensors, limiting platform flexibility and increasing computational cost. Alternatively, models like SpatialVLA~\cite{spatialvla}, BridgeVLA~\cite{li2025bridgevla}, Evo-0~\cite{lin2025evo} and GeoAware-VLA~\cite{geoaware} infer 3D structures from 2D perception but suffer from projection bias, geometric inconsistencies, and occlusion errors. Thus, an efficient framework for 3D spatial perception and spatiotemporal coherence is urgently needed.

%-------------------------------------------------------------------------
% \vspace{-10pt}
\paragraph{World Model-based VLA Models.}
These models~\cite{zhang2024pivot, hu2024video, cen2025worldvla, zhu2025unified, chen2025world4omni, tian2024predictive, assran2025v} jointly predict future visual states and actions. WorldVLA~\cite{cen2025worldvla} unifies action and image through autoregressive prediction, while World4Omni~\cite{chen2025world4omni} employs a pre-trained world model for task-agnostic manipulation. V-JEPA 2~\cite{assran2025v} builds an action-conditioned latent world model from large-scale video pretraining. However, these approaches primarily predict future images rather than achieving true 3D spatial understanding or 4D reasoning. 
% Extending this paradigm toward spatiotemporally consistent modeling remains an open challenge for next-generation VLA systems.

%% file: sec/3_pre.tex
% \vspace{-5pt}
\section{Preliminary \& Problem Definition}

\paragraph{Semantic, Geometric, and Spatial Perception.} For efficient and comprehensive 3D scene encoding, $\mathbf{z}^{\text{sem}}$, $\mathbf{z}^{\text{geo}}$, and $\mathbf{z}^{\text{3D}}$ are derived from visual encoders specialized in semantics, geometry, and spatial perception.

\begin{figure}[t]
  \centering
   \includegraphics[width=1.0\linewidth]{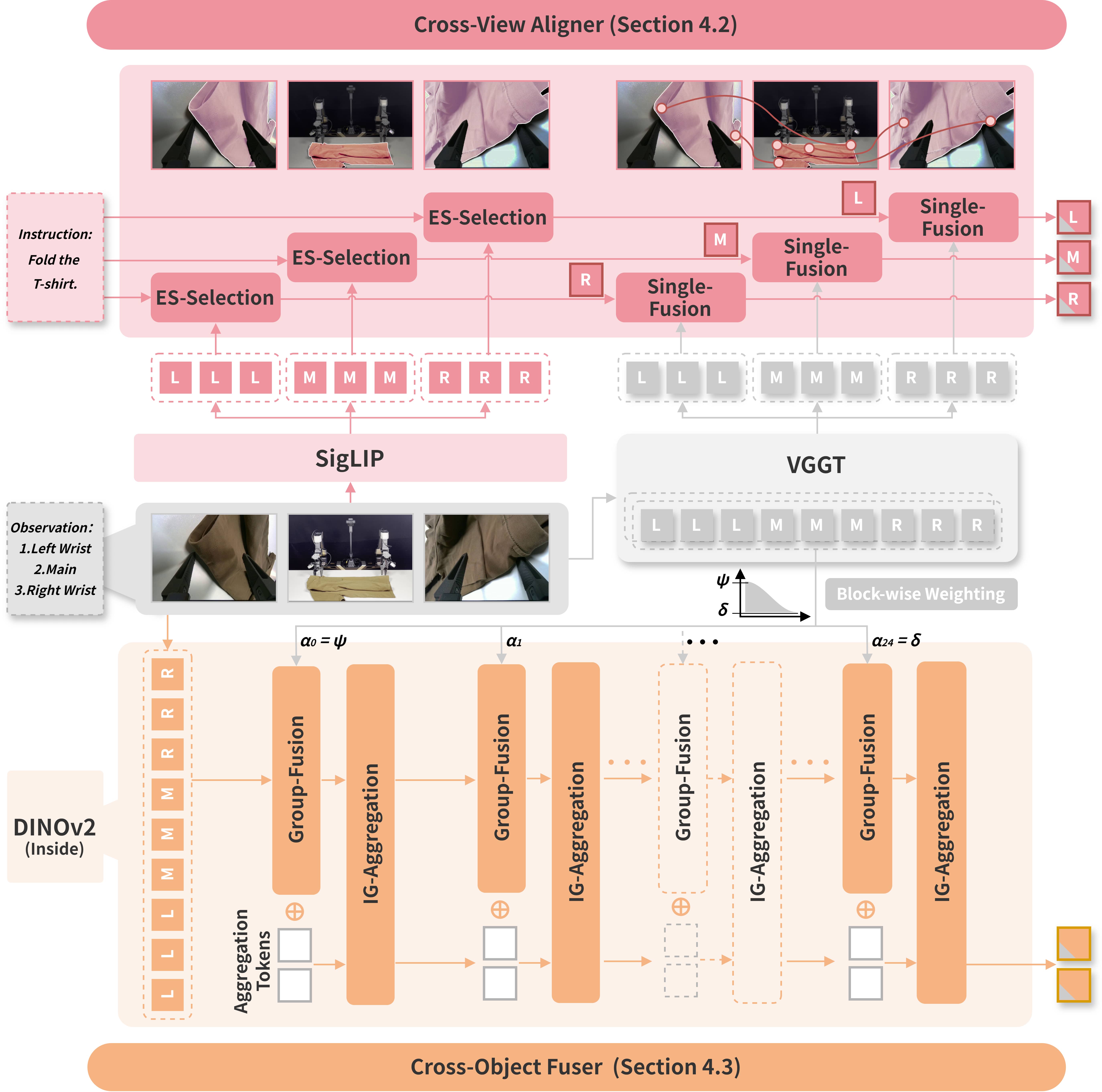}
   % \vspace{-1.8em}
   \caption{\textbf{Efficient 3D-Perception} ensures spatial consistency through the \textcolor[HTML]{EF949E}{Cross-View Aligner (red)} and \textcolor[HTML]{F5B482}{Cross-Object Fuser (orange)}. The former employs an Explicit Semantic Object Selection combined with a frame-wise Single-Fusion strategy, while the latter utilizes Implicit Geometric Relation Aggregation with a multi-frame Group-Fusion strategy to achieve Cross-View Object Semantic Consistency and Cross-Object Spatial Geometric Consistency in the spatial domain. 
   % These representations serve as the spatial perception input for the subsequent 4D-Reasoning phase.
   }
   % \vspace{-12pt}
   \label{fig:3d}
\end{figure}

As shown in Fig.~\ref{fig:3d}, this establishes the foundation for feature selection and fusion design: 
% \textbf{1)} SigLIP~\cite{siglip} is built upon a visual encoder $f_v^{\text{SigLip}}(\cdot)$ and a text encoder $f_t^{\text{SigLip}}(\cdot)$, achieving explicit cross-modal alignment through a sample-wise matching loss. Each visual token $z^{\text{sem}, j}$ within the feature map $z^{\text{sem}} = f_v^{\text{SigLip}}(x)$ thus inherits a degree of linguistic semantics. 
\textbf{1)} SigLIP~\cite{siglip} employs visual and text encoders $f_v^{\text{SigLIP}}(\cdot)$ and $f_t^{\text{SigLIP}}(\cdot)$, aligning modalities via sample-wise matching loss, enabling each visual token $\mathbf{z}^{\text{sem}, j}$ in $\mathbf{z}^{\text{sem}} = f_v^{\text{SigLIP}}(\mathbf{x})$ to inherit linguistic semantics.
% \textbf{2)} DINOv2~\cite{dinov2} generates student and teacher features for augmented views of the same image, using contrastive loss to align features within the same image and disalign those from different images. This self-supervision enables $z^{\text{geo}} = f_v^{\text{Dinov2}}(x)$ to capture geometric consistency. 
\textbf{2)} DINOv2~\cite{dinov2} aligns augmented different views of the same image using contrastive loss, allowing $\mathbf{z}^{\text{geo}} = f_v^{\text{DINOv2}}(\mathbf{x})$ to capture geometric consistency.
% \textbf{3)} VGGT~\cite{wang2025vggt} processes a sequence of $M$ RGB images $ (x_i)_{i=1}^M $ from the same 3D scene, mapping it to a set of 3D annotations: depth maps $ D_i \in \mathbb{R}^{H \times W} $, point maps $ P_i \in \mathbb{R}^{3 \times H \times W} $, and feature grids $ G_i \in \mathbb{R}^{C \times H \times W} $ for point tracking, as shown in the following equation:
\textbf{3)} VGGT~\cite{wang2025vggt} takes $M$ RGB images $(\mathbf{x}_i)_{i=1}^M$ from the same 3D scene and outputs depth maps $D_i$, point maps $P_i$, and feature grids $G_i$ (for point tracking), as expressed below:
% \vspace{-2pt}
\begin{equation} 
% \small
\footnotesize
\text{DPT}\left( f^{\text{VGGT}}_v(\mathbf{x}_i)_{i=1}^M\right)=(D_i, P_i, G_i)_{i=1}^M,
\end{equation}
where $\text{DPT}(\cdot)$ refers to Dense Prediction~\cite{dpt} head, and $f^{\text{VGGT}}_v(\cdot)$ extracts latent features, yielding $\mathbf{z}^{\text{3D}} = f_v^{\text{VGGT}}(\mathbf{x})$.
The latent priors of $G_i$ and $D_i$ in $\mathbf{z}^{\text{3D}}$ enable it to combine with $\mathbf{z}^{\text{sem}}$ and $\mathbf{z}^{\text{geo}}$ for local semantic filtering and global geometric relationship modeling in 3D space.

\begin{figure*}[t]
  \centering
   \includegraphics[width=1.0\linewidth]{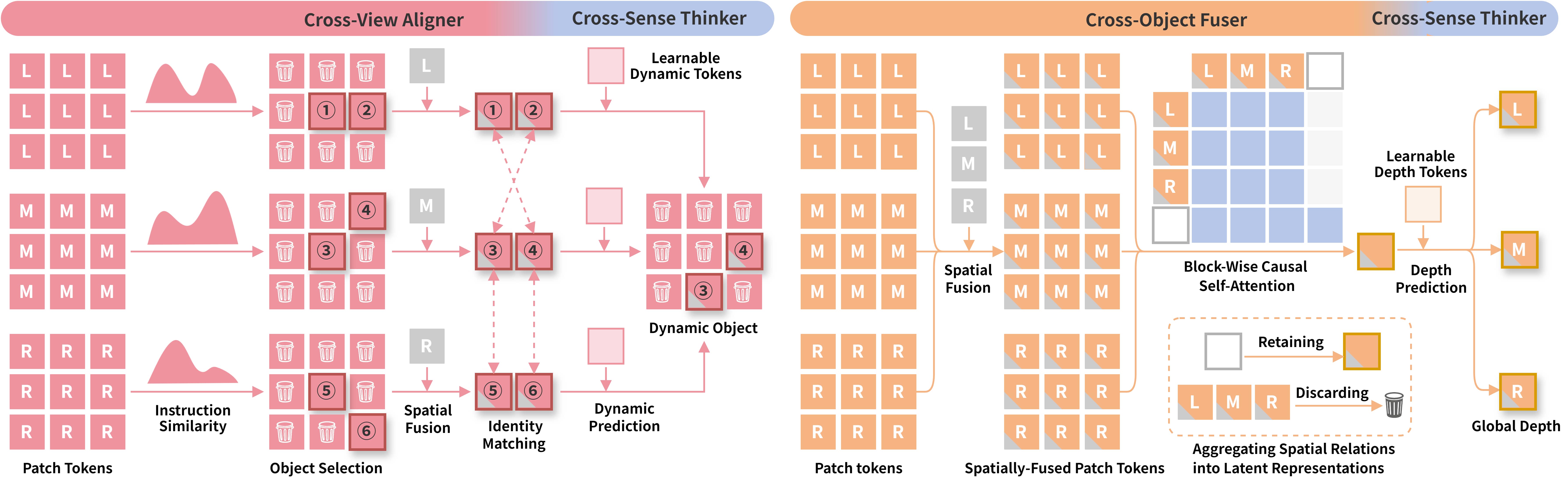}
   % \vspace{-1.8em}
   \caption{\textbf{The Mechanism from 3D-Perception to 4D-Reasoning. }The Cross-View Aligner selects spatial objects with matching identities across different views, and through 4D-Reasoning, further predicts the dynamic object with the same identity from one view to another after an action occurs. The Cross-Object Fuser aggregates global geometric relations to eliminate spatial ambiguity across multiple views, and through 4D-Reasoning, further predicts global depth from the same geometric representation to different views. This progression not only enhances Spatiotemporal Consistency but also reduces visual input, significantly improving training and inference efficiency.}
   % \vspace{-10pt}
   \label{fig:pipline}
\end{figure*}

% \vspace{-10pt} 
\paragraph{Human-Aligned Spatiotemporal Consistency Modeling.} Analogous to how the human retina captures 2D images while the brain integrates spatial cues and predicts future states during manipulation, VLA models should inherit this mechanism, as shown below:

\vspace{-6pt} 
\begin{equation}
\small
\begin{tikzpicture}[>=stealth, node distance=2cm, baseline=(current bounding box.center)]
    \node (2D) {$\textit{2D}$};  
    \node (3D) [right=of 2D] {$\textit{3D}$};  
    \node (4D) [right=of 3D] {$\textit{4D}$}; 
    \draw[->, thick] (2D) -- node[above, font=\footnotesize] {$\textit{construction}$} (3D); 
    \draw[->, thick] (3D) -- node[above, font=\footnotesize] {$\textit{prediction}$} (4D);    
    \draw[->, thick] (4D) -- node[below, font=\footnotesize] {$\textit{refinement}$} (3D);   
\end{tikzpicture}
\end{equation}
\vspace{-6pt}

\noindent The construction of spatially consistent 3D representations from 2D inputs serves as the foundation for spatiotemporal consistency, encompassing: \textbf{1)} the identification and matching of object identities across multiple viewpoints; \textbf{2)} the resolution of visual ambiguities through information complementarity between viewpoints. 
% Meanwhile, VLA models should reason about a dynamic spatial scene rather than merely predicting the next frame, thereby ensuring spatiotemporal consistency for robust and efficient robotic manipulation.
Meanwhile, VLA models should reason about dynamic spatial scenes (rather than merely frames), with the inferred spatial scenes enhancing temporal stability, thereby ensuring spatiotemporal consistency for robust and efficient robotic manipulation.

% Constructing 3D representations from 2D inputs ensures spatiotemporal consistency by \textbf{1)} identifying and matching object identities across viewpoints, and \textbf{2)} resolving visual ambiguities. Additionally, VLA models should reason about a dynamic spatial scene rather than merely predicting the next frame, ensuring spatiotemporal consistency for generalizable and efficient robotic manipulation.

%% file: sec/4_consis.tex
% \vspace{-5pt} 
\section{ConsisVLA-4D}
% \vspace{-2pt} 
\subsection{Proposed Framework}
% \vspace{-2pt} 
\paragraph{Efficient 3D-Perception Phase.}   
As shown in Fig.~\ref{fig:3d} and Fig.~\ref{fig:pipline} left, multi-view observations $\mathcal{I} =\{M, L, R\}$ (Main, Left, Right) are encoded by SigLIP into semantic representations $\mathbf{z}_M^{\text{sem}}$, $\mathbf{z}_R^{\text{sem}}$, and $\mathbf{z}_L^{\text{sem}}$. Guided by the instruction $\mathbf{t}$, these representations are filtered via Explicit Semantic Object Selection $f_{\text{ES-S}}(\cdot)$ and fused frame-wise with the 3D features extracted by VGGT through Single-Fusion $f_{\text{SF}}(\cdot)$, yielding instruction-relevant object representations:
\begin{equation}
\small
\mathbf{z}_{\{M, L, R\}}^{\text{obj-3D}} = f_{\text{SF}}(f_{\text{ES-S}}(\mathbf{z}_{\{M, L, R\}}^{\text{sem}}, \mathbf{t}), \mathbf{z}_{\{M, L, R\}}^{\text{3D}}).
\end{equation}

During spatial geometric feature extraction, single-view depth estimation suffers from scale ambiguity. To address this, tri-view features are integrated for depth completion at each encoder layer. In the $l$-th block of DINOv2 and VGGT, the unified tri-view geometric feature $\mathbf{z}_l^{\text{geo}}$ is fused with the cosine-decayed weighted spatial feature $\mathbf{z}_l^{\text{3D}}$ via Group-Fusion $f_{\text{GF}}(\cdot)$, followed by Implicit Geometric Relation Aggregation $f_{\text{IG-A}}(\cdot)$ to capture spatial relations:
\begin{equation}
\small
(\mathbf{z}_{l+1}^{\text{geo-3D}}, \mathbf{z}_{l+1}^{\text{agg-3D}}) = f_{\text{IG-A}}(f_{\text{GF}}(\mathbf{z}_{l}^{\text{geo}}, \mathbf{z}_l^{\text{3D}}), \mathbf{z}_l^{\text{agg-3D}}).
\end{equation}

% \vspace{-12pt}
\paragraph{Efficient 4D-Reasoning Phase.}  
We further extend 3D-Perception to 4D-Reasoning to expand the spatial consistency into the spatiotemporal domain (Fig.~\ref{fig:pipline} right and Fig.~\ref{fig:4d}). 
On one hand, although objects with the same identity may exhibit appearance differences across different viewpoints, reasoning the object's identity from one perspective to another is crucial for establishing cross-view identity consistency. Furthermore, we focus the reasoning on objects that change as actions unfold, resulting in:
% \vspace{-2pt}
\begin{equation} 
\small
\hat{\mathbf{z}}_{M}^{\text{dyn-4D}} = \text{4D-Reasoning}(\mathbf{z}_{\{M, L, R\}}^{\text{obj-3D}},  \mathbf{t}, \mathbf{0}_{\{M,L,R\}}^{\text{dyn-4D}}).
\end{equation}
On the other hand, we use the aggregated geometric relation $\mathbf{z}_{\mathcal{L}'}^{\text{agg-3D}} $ to infer the depth representations of future multi-view perspectives as actions unfold:
\begin{equation} 
\small
\hat{\mathbf{z}}_{\{M,L,R\}}^{\text{dep-4D}} = \text{4D-Reasoning}(\mathbf{z}_{\mathcal{L}'}^{\text{agg-3D}},  \mathbf{t}, \mathbf{0}^{\text{dep-4D}}),
\end{equation}
where $ \mathbf{0}_{\{M,L,R\}}^{\text{dyn-4D}} $ and $ \mathbf{0}^{\text{dep-4D}} $ denote initialized dynamic and depth tokens, and $\mathcal{L}'$ is the number of encoder layers. 
In comparison, $\mathbf{z}_{M}^{\text{dyn-4D}}$ models future local dynamics through explicit object features, while $ \mathbf{z}_{\{M,L,R\}}^{\text{dep-4D}}$ infers future global depth through implicit spatial relations. We concatenate the initialized action chunk $ \mathbf{0}^{A} $ at the end of the sequence and apply our proposed Spatiotemporal Consistency Attention (SC-Attn) to ultimately obtain the action output:
\begin{equation} 
\footnotesize 
\hat{\mathbf{A}}= \text{SC-Attn}(\mathbf{z}_{\{M, L, R\}}^{\text{obj-3D}}, \mathbf{z}_{\mathcal{L}'}^{\text{agg-3D}}, \mathbf{t}, \mathbf{0}_{\{M,L,R\}}^{\text{dyn-4D}}, \mathbf{0}^{\text{dep-4D}}, \mathbf{0}^A).
\end{equation}

\subsection{Cross-View Object Semantic Consistency}
In the Explicit Semantic Object Selection (ES-Selection) module, layer-wise FiLM~\cite{perez2018film} modulation is applied to further enhance the semantic correlation between $\mathbf{z}_{i,l}^{\text{sem}}$ extracted by the SigLIP visual encoder from the $i$-th viewpoint ($i \in \mathcal{I}$) and $\mathbf{t}  \in \mathbb{R}^{d_t}$ obtained from its paired text encoder:
\begin{equation}
\begin{aligned}
\tilde{\mathbf{z}}_{i,l}^{\text{sem}} = (\mathbf{1}+\gamma(\mathbf{t})) \odot \text{Self-Attn}(\mathbf{z}_{i,l}^{\text{sem}}) + \beta(\mathbf{t}).
\end{aligned}
\end{equation}
Subsequently, the FiLM-modulated final-layer representation $\mathbf{z}_i^{\text{sem}}=\tilde{\mathbf{z}}_{i,\mathcal{L}}^{\text{sem}}$ is utilized to compute the semantic similarity $s_{i,j}$ between each visual token in $\mathbf{z}_i^{\text{sem}}$ and the instruction embedding $\mathbf{t}$, which is defined as:
\begin{equation}
\small
    \mathbf{z}_i^{\text{sem}} = [\mathbf{z}_i^{\text{sem},1}, \mathbf{z}_i^{\text{sem},2}, \ldots, \mathbf{z}_i^{\text{sem},N_i}] \in \mathbb{R}^{N_i \times d_v},
\end{equation}
\vspace{-20pt}
\begin{equation}
\small
    s_{i,j} = \operatorname{sim}(\mathbf{z}_i^{\text{sem},j}, \mathbf{W_t \cdot t}) = \frac{\mathbf{z}_i^{\text{sem},j} (\mathbf{W_t \cdot t}^\top)}{||\mathbf{z}_i^{\text{sem},j}||_2 \cdot||\mathbf{W_t \cdot t}||_2},
\end{equation}
where $\operatorname{sim}(\cdot,\cdot)$ denotes cosine similarity, and $\mathbf{W_t}$ is the mapping matrix from text to viewpoint dimensions. To achieve explicit semantic object selection, we retain only the Top-K tokens most relevant to the instruction:
\begin{equation}
    \small
    \mathcal{S}_i = \operatorname{Top-K}(\{s_{i,1}, s_{i,2}, \ldots, s_{i,N_i}\}, K),
\end{equation}
where $\mathcal{S}_i$ represents the set of selected token indices for the $i$-th viewpoint. The default $K$ is 32, effectively removing substantial visual redundancy. The set of tokens indexed within $\mathbf{z}_i^{\text{sem}}$ is the selected object tokens $\mathbf{z}_i^{\text{obj}}$:
\begin{equation}
\small
\mathbf{z}_i^{\text{obj}} = {\mathbf{z}_i^{\text{sem},j} \mid _{j \in \mathcal{S}_i}} = f_{\text{ES-S}}(\mathbf{z}_{i}^{\text{sem}}, \mathbf{t}).
\end{equation}

% To inject 3D spatial information into $\mathbf{z}_i^{\text{obj}}$ and establish associations between objects with the same identity across different viewpoints, we use Single-Fusion module to integrate $\mathbf{z}_i^{\text{3D}}$ from VGGT frame by frame. Specifically, $\mathbf{z}_i^{\text{obj}}$ serves as the query (Q) in the cross-attention mechanism, while $\mathbf{z}_i^{\text{3D}}$ is used as both the key (K) and value (V):
To inject 3D information into $\mathbf{z}_i^{\text{obj}}$ and establish associations between objects with the same identity across different viewpoints, we introduce Single-Fusion, which performs frame-wise integration of $\mathbf{z}_i^{\text{3D}}$ extracted from VGGT. 
We append $N$ transformer layers, where $\mathbf{z}_i^{\text{obj}}$ serves as the query (Q), and $\mathbf{z}_i^{\text{3D}}$ serves as both the key (K) and value (V). The output of the final layer is taken as $\mathbf{z}_i^{\text{obj-3D}}$:
% \begin{equation}
% \small
% \mathbf{z}_i^{\text{obj-3D}} = \text{softmax}\left(\frac{\mathbf{z}_i^{\text{obj}} (\mathbf{z}_i^{\text{3D}})^\top}{\sqrt{d_k}}\right) \mathbf{z}_i^{\text{3D}}=f
% _{\text{SF}}(\mathbf{z}_i^{\text{obj}}, \mathbf{z}_i^{\text{3D}}).
% \end{equation}
% \vspace{-3pt}
\begin{equation}
% \small
\footnotesize
\begin{aligned}
\mathbf{z}_i^{\text{obj-3D}}
&= \Big( \text{FFN}(\text{Cross-Attn}(\mathbf{z}_i^{\text{obj}}, \mathbf{z}_i^{\text{3D}})) + \text{Res}(\mathbf{z}_i^{\text{obj}}) \Big)\big|_{\text{Layer=1,…,N}} \\
&= f_{\text{SF}}(\mathbf{z}_i^{\text{obj}}, \mathbf{z}_i^{\text{3D}}),
\end{aligned}
\end{equation}
where $\text{Res}(\cdot)$ denotes the residual connection. This leverages VGGT's pre-trained ability of point tracking ($G_i$) from the $i$-th viewpoint and achieves cross-view object semantic consistency using only \textbf{1/8} of the original visual tokens. 

% \begin{equation}
% \small
% \mathbf{z}_i^{\text{obj-3D}} = (\text{FFN}(\text{Cross-Attn}(\mathbf{z}_i^{\text{obj}}, \mathbf{z}_i^{\text{3D}})) + \mathbf{z}_i^{\text{obj}}) |_{\text{Last Layer}}=f
% _{\text{SF}}(\mathbf{z}_i^{\text{obj}}, \mathbf{z}_i^{\text{3D}}).
% \end{equation}

\subsection{Cross-Object Spatial Geometric Consistency}

Due to geometric ambiguity in object positions from a single viewpoint, we jointly process multiple viewpoints (M, L, and R) to abstract the spatial geometric relationships between them. We achieve concise spatial geometric relation aggregation through in-depth feature fusion within the encoder of DINOv2 and VGGT, performed block by block. For the $l$-th block, the weighted 3D representation $\mathbf{z}_l^{\text{3D}}$ is aggregated element-wise with the weight $\alpha_l$ as follows:
\begin{equation}
\small
\mathbf{z}_{l}^{\text{geo-3D}} = (1-{\alpha_l}) \odot \mathbf{z}_{l}^{\text{geo}} + {\alpha_l} \odot \mathbf{z}_l^{\text{3D}} = f_{\text{GF}}(\mathbf{z}_{l}^{\text{geo}}, \mathbf{z}_l^{\text{3D}}).
\end{equation}
The value of $\alpha_l$ is controlled by the layer number $l$, decreasing cosinusoidally with increasing layer depth to maintain a deep yet stable adjustment. The formula is given by:
\begin{equation}
\small
\alpha_l = \psi \cdot \left( \delta + (1-\delta) \cdot \frac{1 + \cos\!\left( \frac{l\pi}{\mathcal{L}'} \right)}{2} \right).
\label{eq:alpha_cos_shift}
\end{equation}
At the 0-th layer, the maximum weight is achieved, $\alpha_0 = \psi$, and at the maximum layer $\mathcal{L}'$, the minimum weight is $\alpha_{\mathcal{L}'} = \delta$. 
After concatenating the learnable $\mathbf{z}_l^{\text{agg-3D}}$ (with $\mathbf{z}_{l+1}^{\text{agg-3D}} = \mathbf{z}_l^{\text{agg}}$ at $l=0$), the combined result is input to the Implicit Geometric Relation Aggregation (IG-Aggregation): 
\begin{equation}
\small
\begin{aligned}
(\mathbf{z}_{l+1}^{\text{geo-3D}}, \mathbf{z}_{l+1}^{\text{agg-3D}}) &= \text{BC-Attn}(\mathbf{z}_{l}^{\text{geo-3D}} \oplus \mathbf{z}_l^{\text{agg-3D}}) \\
&= f_{\text{IG-A}}(\mathbf{z}_{l}^{\text{geo-3D}}, \mathbf{z}_l^{\text{agg-3D}}),
\end{aligned}
\end{equation}
where $\oplus$ denotes the concatenation of two token sets, and $\text{BC-Attn}(\cdot)$ represents block-wise causal self-attention, where causal attention is applied between $\mathbf{z}_{l}^{\text{geo-3D}}$ and $\mathbf{z}_l^{\text{agg-3D}}$, and bidirectional attention is applied within each.

Finally, we use only the final layer output, $\mathbf{z}_{\mathcal{L}'}^{\text{agg-3D}}$. This implicit modeling avoids redundancy with semantic tokens and efficiently aggregates spatial geometric information to maintain cross-object spatial geometric consistency, using only \textbf{1/12-1/8} of the original visual tokens.

\begin{figure}[t]
  \centering
   \includegraphics[width=1.0\linewidth]{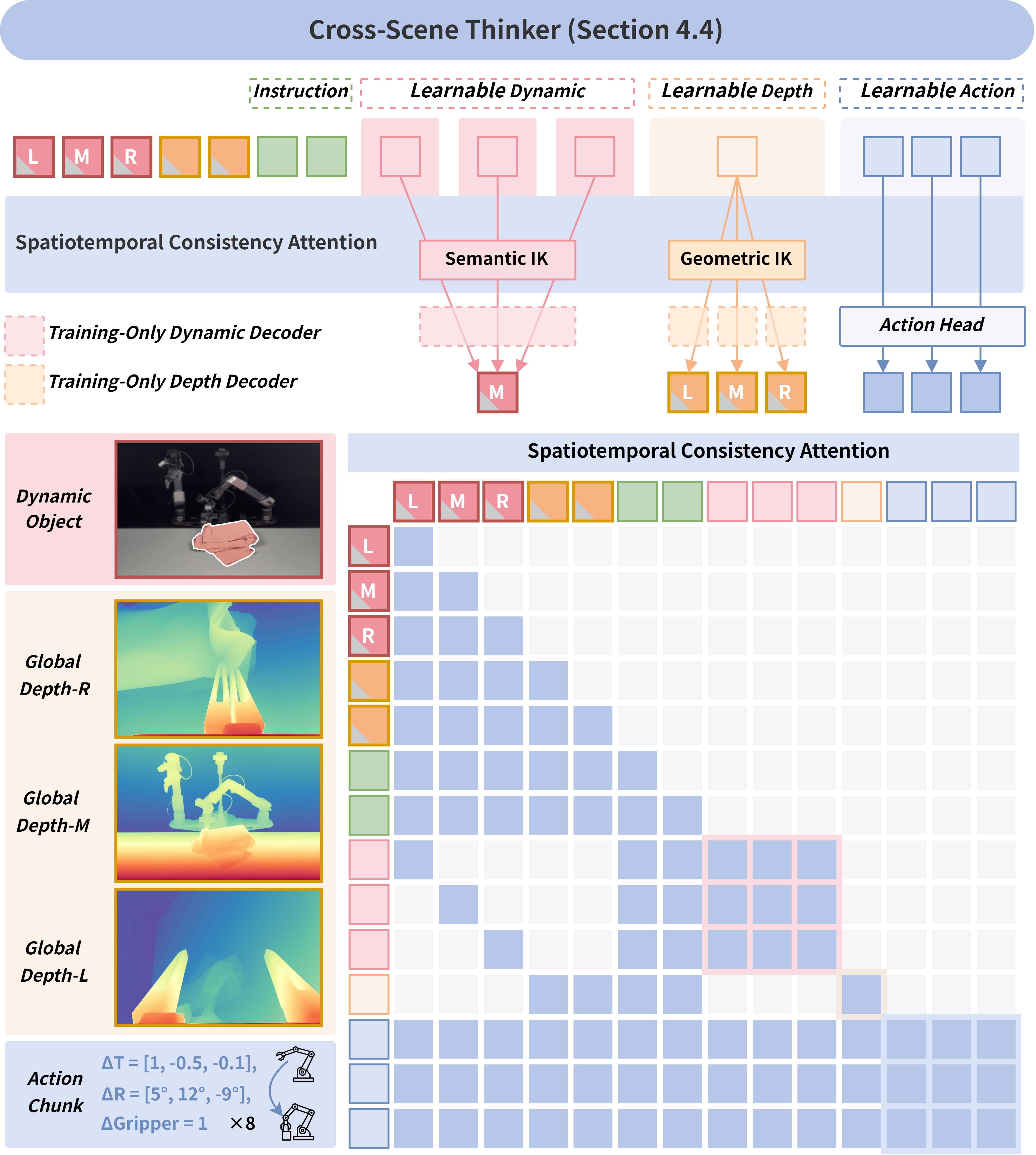}
   % \vspace{-1.8em}
   \caption{\textbf{Efficient 4D-Reasoning.} IK (implicit knowledge). \textcolor[HTML]{ACBCDE}{Cross-Scene Thinker} with Spatiotemporal Consistency Attention (SC-Attn) ensures: \textbf{1)} Three sets of initialized dynamic tokens decode dynamic object representations for one view (CoTracker~\cite{karaev2024cotracker, karaev24cotracker3} supervision), guided by object features from different views; \textbf{2)} One set of initialized depth tokens decodes global depth for three views (Depth-Anything~\cite{yang2024depth1, yang2024depth2} supervision), guided by multi-view geometric relations; \textbf{3)} These predictions serve as intermediate visual reasoning for parallel action decoding.}
   % \vspace{-12pt}
   \label{fig:4d}
\end{figure}

\subsection{Cross-Scene Spatiotemporal Consistency}

As actions unfold, spatial scenes continuously change, extending the requirement of spatial consistency within a single scene to spatiotemporal consistency across scenes. The introduction of the temporal dimension also transforms the perception stage into a reasoning phase toward future states. 

To this end, we design the Cross-Scene Thinker (as shown in Fig.~\ref{fig:4d}) with Spatiotemporal Consistency Attention (SC-Attn), which \textbf{1)} learns implicit semantic and geometric knowledge during training to enable efficient visual reasoning at inference, and \textbf{2)} extends both semantic and geometric consistency into the spatiotemporal domain, thereby enhancing the temporal stability of 3D perception. 

% \vspace{-10pt}
\paragraph{Multi-View Objects to Single-View Dynamic Objects.}
For each viewpoint $i \in \mathcal{I}$, a corresponding learnable token set $\mathbf{0}_{i}^{\text{dyn-4D}}$ is initialized, forming $|\mathcal{I}|$ groups in total. In the SC-Attn module, each dynamic token $\mathbf{0}_{i}^{\text{dyn-4D}}$ is independently guided by its corresponding object representation $\mathbf{z}_{i}^{\text{obj-3D}}$ and the instruction embedding $\mathbf{t}$:
% \begin{equation}
% \small
% \forall i \in \mathcal{I}, \quad \mathbf{0}_{i}^{\text{dyn-4D}} \leftarrow (\mathbf{z}_{i}^{\text{obj-3D}}, \mathbf{t}).
% \label{eq:dyn4d}
% \end{equation}
\begin{equation}
\small
\forall i \in \mathcal{I}, \quad 
\mathbf{0}_{i}^{\text{dyn-4D}}
\xleftarrow{\text{guide}}
(\mathbf{z}_{i}^{\text{obj-3D}}, \mathbf{t}).
\label{eq:dyn4d}
\end{equation}
To enhance spatiotemporal consistency of object dynamics across scenes, the model predicts the dynamic object $\hat{\mathbf{z}}_{i^*}^{\text{dyn-4D}}$ corresponding to a fixed viewpoint $i^* \in \mathcal{I}$ after the action occurs. The dynamic-object loss is formulated as: 
\begin{equation}
\small
\mathcal{L}_{\text{dyn-4D}}
= \big\|
(\hat{\mathbf{z}}_{i^*}^{\text{dyn-4D}} \odot \mathbf{m}_{i^*}^{\text{obj-3D}}) - ({\mathbf{z}}_{i^*}^{\text{dyn-4D}} \odot \mathbf{m}_{i^*}^{\text{obj-3D}})
\big\|_2^2,
\end{equation}
where $\odot$ denotes element-wise multiplication, and $\mathbf{m}_{i^*}^{\text{obj-3D}}$ is the mask that localizes the object position in $\mathbf{z}_{i^*}^{\text{obj-3D}}$.

% \vspace{-10pt}
\paragraph{Abstract Relation to Concrete Global Depths.}
$\mathbf{z}_{\mathcal{L}'}^{\text{agg-3D}}$ aggregates implicit abstract geometric relations across multiple views. In the SC-Attn module, we initialize a set of learnable tokens $\mathbf{0}^{\text{dep-4D}}$ and these tokens are guided jointly by $\mathbf{z}_{\mathcal{L}'}^{\text{agg-3D}}$ and the instruction embedding $\mathbf{t}$, while remaining isolated from the learning processes of $\mathbf{z}_{\mathcal{I}}^{\text{obj-3D}}$ and $\mathbf{0}_{\mathcal{I}}^{\text{dyn-4D}}$:
\begin{equation}
\small
\mathbf{0}^{\text{dep-4D}} \xleftarrow{\text{guide}} (\mathbf{z}_{\mathcal{L}'}^{\text{agg-3D}}, \mathbf{t}).
\label{eq:dep4d}
\end{equation}
Given the spatial ambiguities inherent to single-view perception, the aggregated cross-view geometric relations are decoded into global depth features $\hat{\mathbf{z}}_{\mathcal{I}}^{\text{dep-4D}}$ for each viewpoint. The global-depth loss is defined as:
\begin{equation} 
\small 
\mathcal{L}_{\text{dep-4D}} = \sum_{i=1}^{N_{i}} \mathcal{L}_{\text{dep-4D},i} = \sum_{i=1}^{N_{i}} \big\| \hat{\mathbf{z}}_{i}^{\text{dep-4D}} - {\mathbf{z}}_{i}^{\text{dep-4D}} \big\|_2^2.
\label{eq:20}
\end{equation}
During training, CS-Thinker learns the complete processes from Eq.~\ref{eq:dyn4d} to Eq.~\ref{eq:20} to enhance spatiotemporal consistency understanding. During inference, it operates without explicitly generating dynamic objects or global depth, relying instead on the learned implicit semantic and geometric knowledge for efficient and stable action generation.

% \vspace{-10pt}
\paragraph{Parallel Decoding of Action Chunks in SC-Attn.}
% The prediction of $\hat{z}_{i^*}^{\text{dyn-4D}}$ and $\hat{z}_{\mathcal{I}}^{\text{dep-4D}}$ occurs within the same contextual window as action prediction, serving as comprehensive intermediate visual reasoning for action generation. We initialize a set of learnable action tokens $0^{A}$, guided jointly by the visual reasoning representations and the instruction embedding $\mathbf{t}$. During parallel decoding, the predicted action $\hat{A}$ is optimized using an L1 loss $\mathcal{L}_{\text{action}}$. The overall training objective is formulated as:
The predictions of $\hat{\mathbf{z}}_{i^*}^{\text{dyn-4D}}$ and $\hat{\mathbf{z}}_{\mathcal{I}}^{\text{dep-4D}}$ occur within the same contextual window as action prediction, serving as comprehensive intermediate visual reasoning for action generation. During this process, the initialized action tokens $\mathbf{0}^{A}$ are decoded in parallel to yield $\hat{\mathbf{A}}$, optimized using the L1 loss $\mathcal{L}_{\text{action}}$. The overall training objective is formulated as:
% \begin{equation}
% \small
% \mathcal{L}_{\text{total}}
% = \mathcal{L}_{\text{action}} +  \lambda_{\text{dyn-4D}} \mathcal{L}_{\text{dyn-4D}} + \lambda_{\text{dep-4D}} \mathcal{L}_{\text{dep-4D}}.
% \end{equation}
\begin{equation}
\small
\mathcal{L}_{\text{total}}
= \mathcal{L}_{\text{action}} +  \mathcal{L}_{\text{dyn-4D}} + \mathcal{L}_{\text{dep-4D}}.
\end{equation}

%% file: sec/5_exp.tex
\begin{table}[t]
\caption{\textbf{Simulation Results on LIBERO.} Task success rates across four suites and their overall average.}
    \label{tab:libero}
    % \vspace{-5pt}
    \centering
    % \small
    \footnotesize
    % \scriptsize
    % \tiny
    \setlength{\tabcolsep}{3pt}
    \begin{tabular}{l|cccc|c}
    \toprule
    \textbf{Method} & \textbf{Spatial} & \textbf{Object} & \textbf{Goal} & \textbf{Long} & \textbf{Avg.} \\
    \midrule
     % ACT~\textit{[arXiv'23]}~\cite{ACT}  & 82.0  &  78.8 & 66.1  & 44.0 & 67.7   \\
     Diffusion Policy~\textit{[RSS'23]}~\cite{chi2023diffusion}  & 78.3  &  92.5 & 68.3  & 50.5 & 72.4   \\
     Octo~\textit{[RSS'24]}~\cite{octo}  & 78.9  &  85.7 & 84.6  & 51.1 & 75.1   \\
    OpenVLA~\textit{[CoRL'24]}~\cite{openvla} & 84.7 &  88.4 & 79.2  & 83.7 & 76.5   \\
    OpenVLA-OFT~\textit{[RSS'25]}~\cite{oft} & 97.6  &  98.4 & 97.9  & 94.5 & 97.1   \\
    $\pi_0$~\textit{[RSS'25]}~\cite{pi0} & 96.8  &  98.8 & 95.8  & 85.2 & 94.2   \\
    $\pi_0$-Fast~\textit{[RSS'25]}~\cite{fast} & 96.4  &  96.8 & 88.6  & 60.2 & 85.5   \\
    $\pi_{0.5}$~\textit{[arXiv'25]}~\cite{pi05} &  98.8 &  98.2 &  98.0 & 92.4 &  96.9  \\
    TraceVLA~\textit{[ICLR'25]}~\cite{zheng2024tracevla} & 84.6  &  85.2 & 75.1  & 54.1 & 74.8   \\
     CoT-VLA~\textit{[CVPR'25]}~\cite{cotvla} & 87.5  &  91.6 & 87.6  & 69.0 & 83.9   \\
     % GeoAware-VLA~\textit{[arXiv'25]}~\cite{geoaware} & 95.0  &  100.0 & 99.0  & 93.0 & 96.8   \\
     SpatialVLA~\textit{[RSS'25]}~\cite{spatialvla} & 88.2 &  89.9 & 78.6  & 55.5 & 78.1   \\
    \rowcolor[HTML]{DAE3F5}
    ConsisVLA-4D  & \textbf{98.8}  &  \textbf{99.8} & \textbf{98.0}  & \textbf{95.6} & \textbf{98.1}   \\
    \bottomrule
    \end{tabular}
    % \vspace{-10pt}
\end{table}

\begin{table}[t]
\caption{\textbf{Simulation Results on ManiSkill2.} ``\dag'' denotes results reproduced under identical settings as ConsisVLA-4D.}
    \label{tab:maniskill}
    % \vspace{-5pt}
    \centering
    % \small
    \footnotesize
    % \scriptsize
    % \tiny
    \setlength{\tabcolsep}{3.2pt}
    \begin{tabular}{l|ccc|c}
    \toprule
    \textbf{Method} & \textbf{PickC.} & \textbf{StackC.} & \textbf{PushC.} & \textbf{Avg.} \\
    \midrule
    Octo~\textit{[RSS'24]}~\cite{octo}  & 86\%  &  76\% & -  & 81.0\%  \\
    OpenVLA~\textit{[CoRL'24]}\dag~\cite{openvla} & 67\% & 64\% &  71\% & 67.3\%   \\
   CogACT~\textit{[arXiv'24]}~\cite{li2024cogact} & \textbf{95\%} &  90\% & - & 92.5\%      \\
   GeoVLA~\textit{[arXiv'25]}~\cite{sun2025geovla}  & 90\%  &  90\% & - & 90.0\%     \\
    Dita~\textit{[ICCV'25]}~\cite{hou2025dita} & 79\% &  80\% & - & 79.5\%     \\
    OpenVLA-OFT~\textit{[RSS'25]}\dag~\cite{oft} & 85\% & 93\% & 88\% & 88.7\%   \\
    \rowcolor[HTML]{DAE3F5}
    ConsisVLA-4D  & 93\%  &  \textbf{95\%} & \textbf{95\%} & \textbf{94.3\%}  \\
    \bottomrule
    \end{tabular}
    % \vspace{-10pt}
\end{table}

\begin{table}[t]
\caption{\textbf{Efficiency Optimization Results.} ``\dag'' denotes results reproduced under identical settings as ConsisVLA-4D. \textit{Latency} and \textit{Throughput (T-put)} represent the inference delay and the number of predicted actions per second, while \textit{Cost} indicates the time required for every 10k training steps. The results show that despite the addition of 3D representation input and 4D reasoning processes, ConsisVLA-4D outperforms the most efficient 7B VLA model (OpenVLA-OFT). Ablation on E3D (Efficient 3D-Perception phase) validates its contribution to overall efficiency. 
% All results are evaluated on A800 GPUs, with additional experiments analyzing the impact of hardware configuration on efficiency, as detailed in \textbf{Appx.~C.3}.
}
    \label{tab:efficiency}
    % \vspace{-5pt}
    \centering
    \footnotesize
    \setlength{\tabcolsep}{1.5pt}
    \begin{tabular}{l|cccc}
    \toprule
    \textbf{Method} 
     & \textbf{Latency} $\downarrow$ & \textbf{T-put} $\uparrow$ & \textbf{FLOPs} $\downarrow$ & \textbf{Cost} $\downarrow$  \\
    % \midrule
    \hline
    \rowcolor[HTML]{EFEFEF} \multicolumn{5}{c}{\textit{\scriptsize Simulation: Unimanual tasks}} \\
    \hline
   RT-2-X~\textit{[PMLR'23]}~\cite{Rt-2} &  0.200 s & 5.0 Hz & - & -   \\
   ManipLLM~\textit{[CVPR'24]}~\cite{li2024manipllm} &  - & 2.2 Hz & - & -   \\
   CogACT~\textit{[arXiv'24]}~\cite{li2024cogact}  &  - & 9.8 Hz & - & -   \\
   OpenVLA~\textit{[CoRL'24]}\dag~\cite{openvla} &  0.254 s & 3.9 Hz & 8.48 T & 11.7 h   \\
   TraceVLA~\textit{[ICLR'25]}~\cite{zheng2024tracevla} &  0.192 s & 5.2 Hz & - & -   \\
  SpatialVLA~\textit{[RSS'25]}~\cite{spatialvla} &  0.199 s & 20.1 Hz & - & -   \\
    OpenVLA-OFT~\textit{[RSS'25]}\dag~\cite{oft} & 0.137 s & 58.4 Hz & 8.45 T & 12.3 h   \\
    FiS-VLA~\textit{[NeurIPS'25]}~\cite{oft} & - & 21.9 Hz & - & -   \\
    \rowcolor[HTML]{DAE3F5}
    ConsisVLA-4D  &\textbf{0.110 s} &\textbf{72.7 Hz} &\textbf{4.59 T} &\textbf{8.6 h}  \\
    % \rowcolor[HTML]{EDF1FA}
    % ~~~~w/o CV-Aligner  & s & Hz & 10.46 T & h  \\
    % \rowcolor[HTML]{EDF1FA}
    % ~~~~w/o CO-Fuser   & s & Hz & 10.49 T & h  \\ 
    \rowcolor[HTML]{EDF1FA}
    ~~~~w/o E3D   & 0.204 s & 39.2 Hz & 16.83 T & 22.3 h  \\ 
    \hline
    \rowcolor[HTML]{EFEFEF} \multicolumn{5}{c}{\textit{\scriptsize Real World: Bimanual tasks}} \\
    \hline
   OpenVLA~\textit{[CoRL'24]}\dag~\cite{openvla} & 0.552 s & 1.8 Hz & 16.30 T &  12.8 h   \\
    OpenVLA-OFT~\textit{[RSS'25]}\dag~\cite{oft} & 0.334 s &  74.8 Hz & 14.95 T & 13.7 h   \\
    \rowcolor[HTML]{DAE3F5}
    ConsisVLA-4D  &\textbf{0.231 s} &\textbf{108.2 Hz} &\textbf{9.68 T} &\textbf{ 10.1 h}  \\
    \rowcolor[HTML]{EDF1FA}
    ~~~~w/o E3D   & 0.398 s & 62.8 Hz & 25.70 T & 22.0 h  \\ 
    \bottomrule
    \end{tabular}
    % \vspace{-12pt}
\end{table}

\begin{figure*}[t]
  \centering
   \includegraphics[width=1.0\linewidth]{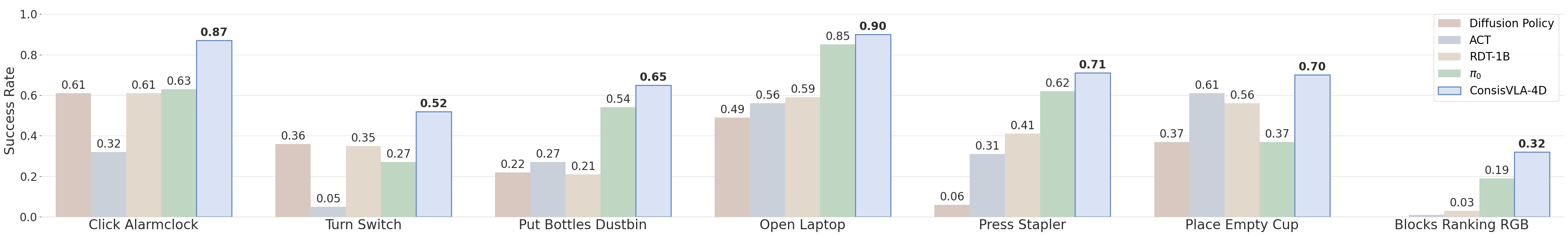}
   % \vspace{-2em}
    \caption{\textbf{Simulation Results on RoboTwin 2.0 Benchmark.} The tasks cover diverse scenarios, with each task conducted in 100 trials.}
   % \vspace{-10pt}
   \label{tab:robotwin}
\end{figure*}

\begin{table*}[t]
\caption{\textbf{Real-World Task Results.} ``\dag'' denotes results reproduced under identical settings as ConsisVLA-4D. The table reports subtask (progressive stage) success rates for four long-horizon tasks on the Galaxea R1 Lite and AgileX Cobot Magic platforms. Decimal values indicate averages over 15 trials, and the average success rate reflects complete task completion.}
    \label{tab:real_world}
    % \vspace{-6pt}
    \centering
    \footnotesize
    \setlength{\tabcolsep}{4.9pt}
    \begin{tabular}{l|ccc|ccc|ccc|ccc|c}
    \toprule
    \textbf{Method} & \multicolumn{3}{c|}{\textbf{Microwave Operation}} & \multicolumn{3}{c|}{\textbf{Banana Peeling}} & \multicolumn{3}{c|}{\textbf{Drawer Arrangement}}  & \multicolumn{3}{c|}{\textbf{T-shirt Folding}} & \textbf{Avg.}\\ 
     & \textbf{Put} & \textbf{+Place}  & \textbf{+Close} & \textbf{Pick}  & \textbf{+Peel} & \textbf{+Place} & \textbf{Pull} & \textbf{+Place} & \textbf{+Push} & \textbf{Step 1} & \textbf{+Step 2} & \textbf{+Step 3} & \\
    % \midrule
    \hline
    \rowcolor[HTML]{EFEFEF} \multicolumn{14}{c}{\textit{\scriptsize Galaxea R1 Lite platform}} \\
    \hline
   OpenVLA\dag~\cite{openvla} & 6.7{\fontsize{6.5pt}{8pt}\selectfont/10} & 5.3{\fontsize{6.5pt}{8pt}\selectfont/10} & 4.7{\fontsize{6.5pt}{8pt}\selectfont/10} & 8.0{\fontsize{6.5pt}{8pt}\selectfont/10} & 0.7{\fontsize{6.5pt}{8pt}\selectfont/10} & 0.7{\fontsize{6.5pt}{8pt}\selectfont/10} & 5.3{\fontsize{6.5pt}{8pt}\selectfont/10} & 3.3{\fontsize{6.5pt}{8pt}\selectfont/10} & 2.0{\fontsize{6.5pt}{8pt}\selectfont/10} & 4.7{\fontsize{6.5pt}{8pt}\selectfont/10} & 4.0{\fontsize{6.5pt}{8pt}\selectfont/10} & 4.0{\fontsize{6.5pt}{8pt}\selectfont/10} & 28.5\%  \\
   SpatialVLA\dag~\cite{spatialvla} & 6.0{\fontsize{6.5pt}{8pt}\selectfont/10} & 5.3{\fontsize{6.5pt}{8pt}\selectfont/10} & 4.7{\fontsize{6.5pt}{8pt}\selectfont/10} & - & - & - & 6.0{\fontsize{6.5pt}{8pt}\selectfont/10} & 3.3{\fontsize{6.5pt}{8pt}\selectfont/10} & 2.7{\fontsize{6.5pt}{8pt}\selectfont/10} & - & - & - & 36.7\% \\
    OpenVLA-OFT\dag~\cite{oft} & 8.0{\fontsize{6.5pt}{8pt}\selectfont/10} & 6.7{\fontsize{6.5pt}{8pt}\selectfont/10} & 6.7{\fontsize{6.5pt}{8pt}\selectfont/10} & 9.3{\fontsize{6.5pt}{8pt}\selectfont/10} & 4.0{\fontsize{6.5pt}{8pt}\selectfont/10} & 3.3{\fontsize{6.5pt}{8pt}\selectfont/10} & 7.3{\fontsize{6.5pt}{8pt}\selectfont/10} & 6.7{\fontsize{6.5pt}{8pt}\selectfont/10} & 6.0{\fontsize{6.5pt}{8pt}\selectfont/10} & 6.7{\fontsize{6.5pt}{8pt}\selectfont/10} & 6.0{\fontsize{6.5pt}{8pt}\selectfont/10} & 4.7{\fontsize{6.5pt}{8pt}\selectfont/10} & 51.8\%  \\
    \rowcolor[HTML]{DAE3F5}
    ConsisVLA-4D  &\textbf{9.3{\fontsize{6.5pt}{8pt}\selectfont/10}} &\textbf{8.7{\fontsize{6.5pt}{8pt}\selectfont/10}} &\textbf{8.0{\fontsize{6.5pt}{8pt}\selectfont/10}} &\textbf{10{\fontsize{6.5pt}{8pt}\selectfont/10}} & \textbf{6.0{\fontsize{6.5pt}{8pt}\selectfont/10}} & \textbf{6.0{\fontsize{6.5pt}{8pt}\selectfont/10}} & \textbf{8.7{\fontsize{6.5pt}{8pt}\selectfont/10}} & \textbf{8.0{\fontsize{6.5pt}{8pt}\selectfont/10}} & \textbf{6.7{\fontsize{6.5pt}{8pt}\selectfont/10}} & \textbf{9.3{\fontsize{6.5pt}{8pt}\selectfont/10}} & \textbf{8.7{\fontsize{6.5pt}{8pt}\selectfont/10}} & \textbf{7.3{\fontsize{6.5pt}{8pt}\selectfont/10}} & \textbf{70.0\%} \\
    \hline
    \rowcolor[HTML]{EFEFEF} \multicolumn{14}{c}{\textit{\scriptsize AgileX Cobot Magic platform}} \\
    \hline
    OpenVLA\dag~\cite{openvla} & 6.0{\fontsize{6.5pt}{8pt}\selectfont/10} & 5.3{\fontsize{6.5pt}{8pt}\selectfont/10} & 4.7{\fontsize{6.5pt}{8pt}\selectfont/10} & 8.0{\fontsize{6.5pt}{8pt}\selectfont/10} & 1.3{\fontsize{6.5pt}{8pt}\selectfont/10} & 0.7{\fontsize{6.5pt}{8pt}\selectfont/10} & 5.3{\fontsize{6.5pt}{8pt}\selectfont/10} & 3.3{\fontsize{6.5pt}{8pt}\selectfont/10} & 2.7{\fontsize{6.5pt}{8pt}\selectfont/10} & 4.7{\fontsize{6.5pt}{8pt}\selectfont/10} & 4.7{\fontsize{6.5pt}{8pt}\selectfont/10} & 4.0{\fontsize{6.5pt}{8pt}\selectfont/10} & 30.0\%  \\
    OpenVLA-OFT\dag~\cite{oft} & 8.0{\fontsize{6.5pt}{8pt}\selectfont/10} & 6.7{\fontsize{6.5pt}{8pt}\selectfont/10} & 6.0{\fontsize{6.5pt}{8pt}\selectfont/10} & 9.3{\fontsize{6.5pt}{8pt}\selectfont/10} & 3.3{\fontsize{6.5pt}{8pt}\selectfont/10} & 2.7{\fontsize{6.5pt}{8pt}\selectfont/10} & 7.3{\fontsize{6.5pt}{8pt}\selectfont/10} & 6.7{\fontsize{6.5pt}{8pt}\selectfont/10} & 6.7{\fontsize{6.5pt}{8pt}\selectfont/10} & 6.7{\fontsize{6.5pt}{8pt}\selectfont/10} & 5.3{\fontsize{6.5pt}{8pt}\selectfont/10} & 4.7{\fontsize{6.5pt}{8pt}\selectfont/10} & 50.3\%  \\
    \rowcolor[HTML]{DAE3F5}
    ConsisVLA-4D  &\textbf{9.3{\fontsize{6.5pt}{8pt}\selectfont/10}} &\textbf{8.7{\fontsize{6.5pt}{8pt}\selectfont/10}} &\textbf{8.0{\fontsize{6.5pt}{8pt}\selectfont/10}} &\textbf{9.3{\fontsize{6.5pt}{8pt}\selectfont/10}} & \textbf{6.0{\fontsize{6.5pt}{8pt}\selectfont/10}} & \textbf{6.0{\fontsize{6.5pt}{8pt}\selectfont/10}} & \textbf{8.7{\fontsize{6.5pt}{8pt}\selectfont/10}} & \textbf{7.3{\fontsize{6.5pt}{8pt}\selectfont/10}} & \textbf{6.0{\fontsize{6.5pt}{8pt}\selectfont/10}} & \textbf{9.3{\fontsize{6.5pt}{8pt}\selectfont/10}} & \textbf{8.7{\fontsize{6.5pt}{8pt}\selectfont/10}} & \textbf{7.3{\fontsize{6.5pt}{8pt}\selectfont/10}} & \textbf{68.3\%} \\
    \bottomrule
    \end{tabular}
    % \vspace{-12pt}
\end{table*}

% \begin{table}[t]
% \caption{\textbf{SimplerEnv-Bridge with WidowX robot.} The results demonstrate that the CLS token proposed by SigLIP, with its rich knowledge and generalization ability, most effectively optimizes the weights between.}
%     \label{tab:ablation_cv+co}
%     \vspace{-5pt}
%     \centering
%     % \small
%     \footnotesize
%     % \scriptsize
%     % \tiny
%     \setlength{\tabcolsep}{2pt}
%     \begin{tabular}{l|cccc|c}
%     \toprule
%     \textbf{Method} & \textbf{Spoon} & \textbf{Carrot} & \textbf{Cube} & \textbf{Eggplant} & \textbf{Avg.} \\
%     \midrule
%      Diffusion Policy~\textit{[RSS'23]}~\cite{chi2023diffusion}  & 4.2  &  0.0 & 0.0  & 0.0 & 1.1   \\
%      Octo~\textit{[RSS'24]}~\cite{octo}   & 15.8 & 12.5 & 0.0 & 41.7 & 17.5 \\
%     OpenVLA~\textit{[CoRL'24]}~\cite{openvla} & 4.2  &  0.0 & 0.0  & 12.5 & 8.4   \\
%     RoboVLMs~\textit{[arXiv'24]}~\cite{oft} & 45.8 & 20.8 & 4.2 & 79.2 & 37.5 \\
%      CogACT~\textit{[arXiv'24]}~\cite{geoaware} & 71.7 & 50.8 & 15.0 & 67.5 & 51.3 \\
%     $\pi_0$~\textit{[RSS'25]}~\cite{pi0} & 63.3 & 58.8 & 21.3 & 79.2 & 55.7 \\
%     TraceVLA~\textit{[ICLR'25]}~\cite{zheng2024tracevla} & 12.5 & 16.6 & 16.6 & 65.0 & 27.7 \\
%      SpatialVLA~\textit{[RSS'25]}~\cite{spatialvla} & 16.7 & 25.0 & 29.2 & 100.0 & 42.7 \\
%     \rowcolor[HTML]{DAE3F5}
%     ConsisVLA-4D  & xx.x  &  xx.x & xx.x  & xx.x & xx.x   \\
%     \bottomrule
%     \end{tabular}
%     \vspace{-10pt}
% \end{table}

% \vspace{-10pt}
\section{Experiments}
% \vspace{-2pt}
\subsection{Experimental Setup}
% \vspace{-2pt}
\paragraph{Implementation Details.} 
Training uses 4× A800 GPUs, and real-world inference runs on a single RTX 5090.

% \vspace{-10pt}
\paragraph{Simulation Benchmark.} 
We conduct evaluations across multiple simulation benchmarks, including:
\textbf{1)} the four task suites of LIBERO~\cite{libero}—Spatial, Object, Goal, and Long;
\textbf{2)} three pick-and-place tasks emphasizing spatial scene perception in ManiSkill2~\cite{gu2023maniskill2}; and
\textbf{3)} diverse ALOHA-based~\cite{aloha} bimanual tasks within RoboTwin 2.0~\cite{chen2025robotwin}.

% \vspace{-11pt}
\paragraph{Real-World Platforms and Tasks.} 
ConsisVLA-4D is deployed on both the AgileX Cobot Magic~\cite{aloha} and Galaxea R1 Lite~\cite{galaxea} platforms for real-world evaluations, covering four categories of long-horizon tasks: \textbf{1)} Microwave Operation, \textbf{2)} Banana Peeling, \textbf{3)} Drawer Arrangement, and \textbf{4)} T-shirt Folding.
Each task includes 60, 60, 60, and 45 human-teleoperated demonstrations across both platforms.

% \vspace{-11pt}
\paragraph{Baselines.} 
In simulation, ConsisVLA-4D is compared with multiple SOTA methods, including OpenVLA~\cite{openvla}, SpatialVLA~\cite{spatialvla}, $\pi_0$~\cite{pi0}, CoT-VLA~\cite{cotvla}, and OpenVLA-OFT~\cite{oft}.
In real-world, we reproduce three representative models, including OpenVLA (base model), and OpenVLA-OFT (the best 7B baseline in performance and efficiency), with training and deployment settings identical.
For efficiency, we conduct fair comparisons between these three baselines with ConsisVLA-4D on the LIBERO and Galaxea R1 Lite platforms.

\begin{figure*}[t]
  \centering
   \includegraphics[width=1.0\linewidth]{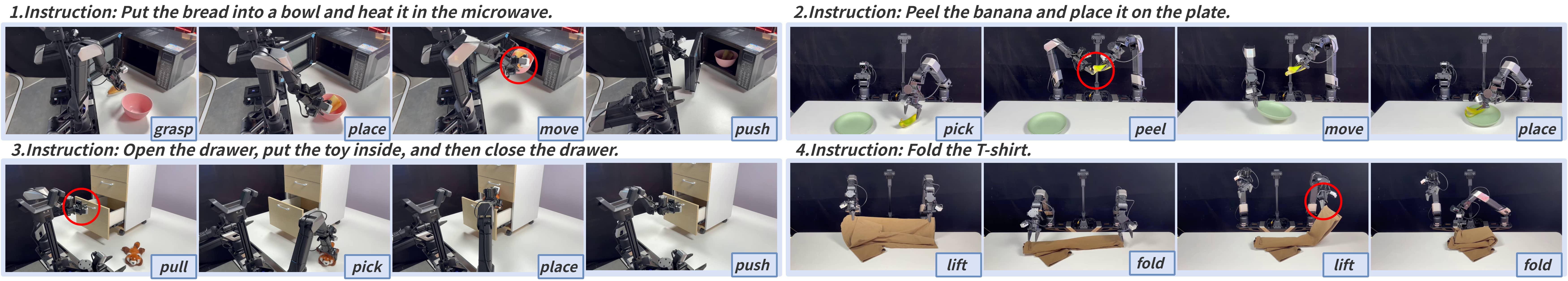}
   % \vspace{-1.8em}
   \caption{Visualization of ConsisVLA-4D performing four long-horizon real-world manipulation tasks on the Galaxea R1 Lite platform, illustrating key execution-stage observations. 
   \textcolor[HTML]{FF0000}{Red circles} highlight fine-grained gripper operations, including \textit{grasping bowl edge}, \textit{peeling banana}, \textit{holding drawer handle}, and \textit{pinching cloth corner}.
   }
   % \vspace{-10pt}
   \label{fig:vis1}
\end{figure*}

\begin{figure*}[t]
  \centering
   \includegraphics[width=1.0\linewidth]{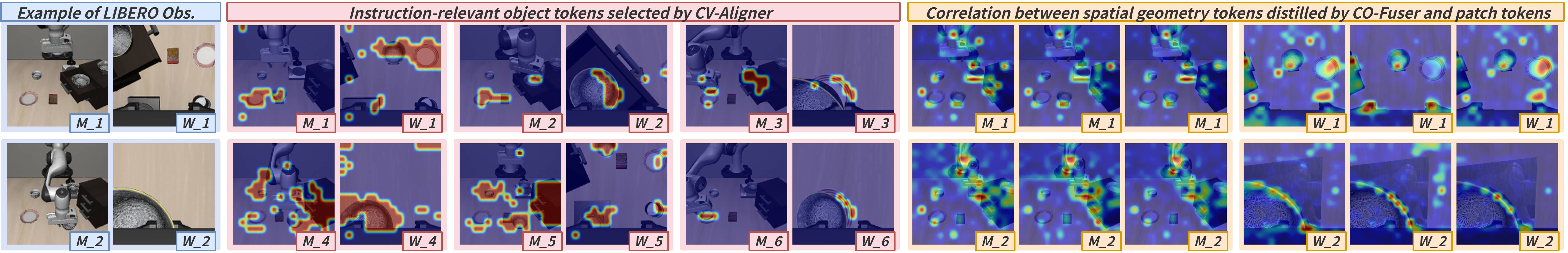}
   % \vspace{-1.8em}
   \caption{Visualization of \textbf{1)} LIBERO observation examples, where \textit{M} and \textit{W} denote main and wrist views; \textbf{2)} instruction-relevant object tokens~\includegraphics[height=0.9em]{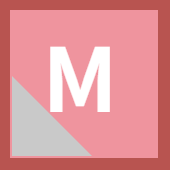}~\includegraphics[height=0.9em]{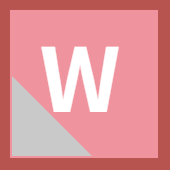} extracted by CV-Aligner, identifying objects such as bowl on/next to the cookie box, bowl in the top drawer of the wooden cabinet, stove, and plate; and \textbf{3)} aggregated spatial geometry tokens~\includegraphics[height=0.9em]{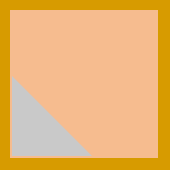} from CO-Fuser capturing geometric relations across multiple views.}
   % \vspace{-12pt}
   \label{fig:vis2}
\end{figure*}

\begin{table}[t]
\caption{\textbf{Ablation Study on CV-Aligner and CO-Fuser.} Ablation components include ES-Selection, Single-Fusion from CV-Aligner, and Group-Fusion, IG-Aggregation from CO-Fuser.}
    \label{tab:ablation_cv+co}
    % \vspace{-6pt}
    \centering
    % \small
    \footnotesize
    % \scriptsize
    % \tiny
    \setlength{\tabcolsep}{3pt}
    \begin{tabular}{cc|cc|cc}
    \toprule
    \multicolumn{2}{c|}{\textbf{CV-Aligner}} & \multicolumn{2}{c|}{\textbf{CO-Fuser}} &  \textbf{LIBERO SR $\uparrow$} & \textbf{Real-World SR $\uparrow$}\\
    ES-Sel. & S-Fus. & G-Fus. & IG-Agg. &  Four Suites & Two Tasks\\
    \midrule
       &  \ding{51} &  \ding{51} & \ding{51}  & 93.9~\textcolor{gray}{(-4.2)} & 71.7~\textcolor{gray}{(-6.6)}   \\
      \ding{51} &   & \ding{51}  & \ding{51}  & 95.5~\textcolor{gray}{(-2.6)} & 73.3~\textcolor{gray}{(-5.0)}   \\
    \ding{51}  & \ding{51} &   & \ding{51}  & 95.6~\textcolor{gray}{(-2.5)} & 70.0~\textcolor{gray}{(-8.3)}  \\
    \ding{51}  & \ding{51} & \ding{51}  &  & 91.7~\textcolor{gray}{(-6.4)} & 68.3~\textcolor{gray}{(-10.0)}   \\
    \rowcolor[HTML]{DAE3F5}
    \ding{51}  & \ding{51} & \ding{51}  & \ding{51} & \textbf{98.1} & \textbf{78.3}  \\
    \bottomrule
    \end{tabular}
    % \vspace{-15pt}
\end{table}

% \vspace{-2pt}
\subsection{Overall Performance \& Efficiency}
% \vspace{-3pt}
\paragraph{Simulation Performance.} \textbf{1)} As illustrated in Tab.~\ref{tab:libero}, ConsisVLA-4D outperforms all methods across the four LIBERO suites. Particularly, it achieves exceptional success rates of \textbf{98.8\%} and \textbf{99.8\%} in the Spatial and Object suites, which assess spatial perception and object recognition, respectively. The average success rate exceeds that of comparative methods SpatialVLA (specialized in spatial modeling) and CoT-VLA (specialized in visual reasoning) by \textbf{20\%} and \textbf{14.2\%}, highlighting the advantages of our 3D spatial representation and 4D visual reasoning. \textbf{2)} As shown in Tab.~\ref{tab:maniskill} and Fig.~\ref{tab:robotwin}, ConsisVLA-4D demonstrates stable superiority across the three classic tasks in ManiSkill2 (PickCube, StackCube, PushCube) and seven diverse tasks in RoboTwin 2.0, testing fine spatial perception and robust bimanual manipulation, respectively.

% \vspace{-12pt}
\paragraph{Simulation Efficiency.}
As illustrated in Tab.~\ref{tab:efficiency}, despite adding approximately 2B parameters (mainly from VGGT), ConsisVLA-4D achieves \textbf{2.31$\times$} and \textbf{1.25$\times$} speedups in inference latency and \textbf{1.36$\times$} and \textbf{1.43$\times$} speedups in training cost compared to the base 7B baseline (OpenVLA) and the most efficient 7B baseline (OpenVLA-OFT), respectively.

% \vspace{-12pt}
\paragraph{Real-World Performance.} 
As shown in Tab.~\ref{tab:real_world}, 15-trial results are normalized to a 10-trial equivalent for intuitiveness. ConsisVLA-4D leads significantly in both phased and final success rates across 4 diverse long-horizon bimanual tasks, with strong performance stably maintained across deployment platforms (\textbf{±1.7\%}). Enhanced spatiotemporal consistency gives ConsisVLA-4D greater advantages in these 4 real-world tasks, where spatial understanding and scenario variations are more emphasized. Notably, its real-world results are nearly consistent with those on RoboTwin 2.0 (ALOHA manipulator), demonstrating robust sim-to-real transfer capability.

% \vspace{-12pt}
\paragraph{Real-World Efficiency.} 
Compared to simulation, the real-world setup adds a new viewpoint input, increases the action chunk size from 8 to 25, and reduces the training batch size from 64 to 32. As shown in Tab.~\ref{tab:efficiency}, ConsisVLA-4D maintains similar efficiency gains in inference latency and training cost as in simulation. The significant throughput improvement by \textbf{33.4} Hz enables real-time, smooth operation of large VLA models on real hardware. When \textbf{E3D} is ablated, the large spatial information representation leads to a significant drop in model efficiency.

\subsection{Ablation Studies}
% \vspace{-3pt}
We conduct ablation experiments in simulation (LIBERO) and real-world (Galaxea R1 Lite). Two tasks (Microwave Operation, T-shirt Folding) are selected, with 15 trials per task extended to 30 to ensure stable ablation results.

% \vspace{-12pt}
\paragraph{CV-Aligner and CO-Fuser.} 
As shown in Tab.~\ref{tab:ablation_cv+co}, for both simulation and real-world results: \textbf{1)} Removing ES-Sel. and S-Fus. from CV-Aligner prevents the filtering of redundant visual inputs and fails to align object identities across different viewpoints, leading to success rate drops of \textbf{7.0\%} and \textbf{10.0\%}. \textbf{2)} Removing G-Fus. and IG-Agg. from CO-Fuser causes visual ambiguities in spatial relationships between objects, resulting in success rate drops of \textbf{8.2\%} and \textbf{13.3\%}. These results demonstrate that the consistency modeling of each module plays a crucial role in ensuring the model's efficient understanding of complex 3D scenes. Moreover, all modules are adaptively designed, and swapping them with counterparts in SigLIP and DINOv2 degrades performance.

% \vspace{-12pt}
\paragraph{CS-Thinker.} 
% As shown in Tab.~\ref{tab:ablation_cs}, ablation of Dyn. O. and Glob. D. in SC-Attn cause success rate drops of \textbf{2.7\%-4.8\%} and \textbf{5.7\%-11.6\%} in simulation and real-world, respectively, highlighting the contribution of both local object and global depth in 4D-Reasoning. 
% Additionally, comparison with causal and bidirectional attention shows that SC-Attn’s targeted modulation enhances the role of each visual element in spatiotemporal consistency modeling, thus improving action execution success rates.
As shown in Tab.~\ref{tab:ablation_cs}, removing Dyn. O. and Glob. D. from SC-Attn causes success rate drops of \textbf{2.7\%-4.8\%} in simulation and \textbf{5.7\%-11.6\%} in the real world. These results highlight the importance of local object and global depth in 4D-Reasoning and action prediction. 
Moreover, comparison with causal and bidirectional attention shows that SC-Attn’s targeted modulation enhances the role of each visual element in spatiotemporal consistency modeling, thus improving action execution success rates.

\begin{table}[t]
\caption{\textbf{Ablation Study on CS-Thinker.} Dyn. O. and Glob. D. represent the training-only dynamic objects and global depth representations in 4D-Reasoning, respectively.}
    \label{tab:ablation_cs}
    % \vspace{-6pt}
    \centering
    % \small
    \footnotesize
    % \scriptsize
    % \tiny
    \setlength{\tabcolsep}{3pt}
    \begin{tabular}{cc|c|cc}
    \toprule
   \textbf{ Dyn. O.} & \textbf{Glob. D.} & \textbf{Attention}  & \textbf{LIBERO SR $\uparrow$} & \textbf{Real-World SR $\uparrow$}  \\
    \midrule
       &   &  SC-Attn &  93.3~\textcolor{gray}{\scriptsize (-4.8)} & 66.7~\textcolor{gray}{\scriptsize (-11.6)}  \\
       & \ding{51}  & SC-Attn  &  95.4~\textcolor{gray}{\scriptsize (-2.7)} & 73.3~\textcolor{gray}{\scriptsize (-5.0)}  \\
      \ding{51} &   &  SC-Attn &  94.7~\textcolor{gray}{\scriptsize (-3.4)} & 71.6~\textcolor{gray}{\scriptsize (-6.7)}  \\
      \ding{51} & \ding{51}  & Causal  &  90.9~\textcolor{gray}{\scriptsize (-7.2)} & 66.7~\textcolor{gray}{\scriptsize (-11.6)}  \\
      \ding{51} & \ding{51}  & Bidirectional  &  92.2~\textcolor{gray}{\scriptsize (-5.9)} & 68.3~\textcolor{gray}{\scriptsize (-10.0)}  \\
      \rowcolor[HTML]{DAE3F5}
      \ding{51} & \ding{51}  & SC-Attn  &  \textbf{98.1} & \textbf{78.3}  \\
    \bottomrule
    \end{tabular}
    % \vspace{-15pt}
\end{table}

\begin{table}[t]
\caption{\textbf{Ablation Study on sparsification ratio (Spf.Ratio).} ``\dag'' denotes reproduced results of FastV and SliME. $\mathbf{0}^{\text{4D}}$ represents the sum of $\mathbf{0}_{\mathcal{I}}^{\text{dyn-4D}}$ and $\mathbf{0}^{\text{dep-4D}}$ token counts.}
% \vspace{-6pt}
    \label{tab:sparsification}
    \centering
    \footnotesize
    % \small
    \setlength{\tabcolsep}{1.4pt}  % 6pt
    \begin{tabular}{c|ccc|cc}
    \toprule
    \textbf{Spf.Ratio} & {$\mathbf{z}_{\mathcal{I}}^{\text{obj-3D}}$}  & {$\mathbf{z}_{\mathcal{L}'}^{\text{agg-3D}}$} & {$\mathbf{0}^{\text{4D}}$} & \textbf{LIBERO SR $\uparrow$} & \textbf{Real-World SR $\uparrow$} \\
    \midrule
    $\approx$ 1/4 & 128 & 128 & 30 & 98.0~\textcolor{gray}{\scriptsize (-0.1)} & 80.0~\textcolor{black}{\scriptsize (+0.7)} \\
  \rowcolor[HTML]{DAE3F5}
    $\approx$ 1/8 & 64 & 64 & 18 & \textbf{98.1} & \textbf{78.3}\\
    $\approx$ 1/16 & 32 & 32 & 12 & 94.9~\textcolor{gray}{\scriptsize (-3.2)} & 68.3~\textcolor{gray}{\scriptsize (-10.0)}\\
    \midrule
    \midrule
     $\approx$ 1/8 &  & FastV\dag~\cite{fastv}  &  & 88.8~\textcolor{gray}{\scriptsize (-9.3)} & 50.0~\textcolor{gray}{\scriptsize (-28.3)}\\
    $\approx$ 1/8 &  & SliME\dag~\cite{slime} &  & 85.6~\textcolor{gray}{\scriptsize (-12.5)} & 46.7~\textcolor{gray}{\scriptsize (-31.6)} \\
    \bottomrule
    \end{tabular}
    % \vspace{-13pt}
\end{table}

% \vspace{-12pt}
\paragraph{Sparsification Ratio.} 
Tab.~\ref{tab:sparsification} presents results under varying sparsification ratios $R \in \{1/4, 1/8, 1/16\}$. The $R = 1/8$ setting achieves a favorable balance between performance and efficiency. Furthermore, compared to other general sparsification methods, such as FastV~\cite{fastv} and SliME~\cite{slime}, at the same 8× compression level, ConsisVLA-4D demonstrates significant advantages.

\subsection{Qualitative Analysis}
% \vspace{-3pt}
Fig.~\ref{fig:vis1} illustrates the model's ability to execute manipulation tasks smoothly and accurately across four long-horizon real-world scenarios. Fig.~\ref{fig:vis2} visualizes: \textbf{1)} selected local object tokens from patch tokens at different viewpoints; and \textbf{2)} the correlation of a global spatial relation aggregation token with all patch tokens across viewpoints. Based on this correlation, ConsisVLA-4D achieves state-of-the-art performance with only 1\textbf{/8} of the visual tokens (see Tab.~\ref{tab:sparsification}).

%% file: sec/6_conclusion.tex
% \vspace{-5pt}
\section{Conclusion}
% \vspace{-5pt}
ConsisVLA-4D advances spatiotemporal reasoning for robotic manipulation by unifying efficient 3D perception and 4D visual reasoning. Through the integration of CV-Aligner, CO-Fuser, and CS-Thinker, it achieves cross-view, cross-object, and cross-scene consistency, enabling robust and efficient understanding of dynamic environments. Experiments demonstrate substantial improvements in both performance and efficiency over prior VLAs. We expect this work to provide useful insights for embodied research.

%% file: sec/X_suppl.tex
\clearpage
\setcounter{page}{1}
\maketitlesupplementary
\renewcommand{\thesection}{\Alph{section}}

\section{Implementation Details}
\label{sec:implementation_details}

\subsection{Model Details}
\paragraph{CV-Aligner.} CV-Aligner aligns multi-view visual observations with task instructions and integrates 3D information from VGGT~\cite{wang2025vggt} through a three-step process: feature modulation, token selection, and cross-view fusion.
\begin{itemize}[leftmargin=*]
\item \textbf{Step 1 (FiLM Modulation).} Although each visual token $\mathbf{z}^{\text{sem}, j}$ from SigLIP~\cite{siglip} inherits pretrained linguistic semantics, we further strengthen the alignment between observations and the task instruction $\mathbf{t}$ for robotic manipulation by modulating SigLIP with FiLM~\cite{perez2018film} scale and shift vectors $\gamma_i(\mathbf{t})$ and $\beta_i(\mathbf{t})$. Specifically, the instruction $\mathbf{t}$ is projected into the visual embedding space at every SigLIP transformer layer to produce layer-specific $\gamma_i(\mathbf{t})$ and $\beta_i(\mathbf{t})$, which are then applied as independent per-token multiplicative and additive modulations.
\item \textbf{Step 2 (ES-Selection).}
We first encode the task instruction $\mathbf{t}$ using $f_t^{\text{SigLIP}}(\cdot)$, then compute its cosine similarity with every visual token from each camera view. For each view, we retain the top $1/8$ of the original tokens (256 $\rightarrow$ 32). ES-Selection is performed independently for each view, producing $\mathbf{z}_{\{M, L, R\}}^{\text{sem}}$.
% \item \textbf{Step 3 (Single-Fusion).} We perform cross-attention between 
% $\mathbf{z}_{\{M, L, R\}}^{\text{sem}}$ and the VGGT-derived 
% $\mathbf{z}_{\{M, L, R\}}^{\text{3D}}$. A 4-layer Transformer~\cite{vaswani2017attention} with a 
% 1152-dimensional hidden size, 16 attention heads, and a 2752-dimensional 
% feed-forward network is employed; visual features (1024-d) are linearly 
% projected to the Transformer hidden space as keys and values. This yields 
% $\mathbf{z}_{\{M, L, R\}}^{\text{obj-3D}}$, integrating VGGT’s 3D information 
% while maintaining 32 tokens per view.
\item \textbf{Step 3 (Single-Fusion).} We fuse $\mathbf{z}_{\{M, L, R\}}^{\text{sem}}$ with the VGGT-derived $\mathbf{z}_{\{M, L, R\}}^{\text{3D}}$ via a cross-attention module. The fusion is instantiated as a 4-layer Transformer~\cite{vaswani2017attention} configured with a 1152-dimensional hidden size, 16 attention heads, and a 2752-dimensional feed-forward block. The 1024-dimensional visual features are linearly projected into the Transformer’s hidden space to serve as keys and values. This process produces $\mathbf{z}_{\{M, L, R\}}^{\text{obj-3D}}$, effectively injecting VGGT’s 3D cues while preserving a fixed budget of 32 tokens per view.
\end{itemize}

\begin{figure}[t]
  \centering
   \includegraphics[width=0.75\linewidth]{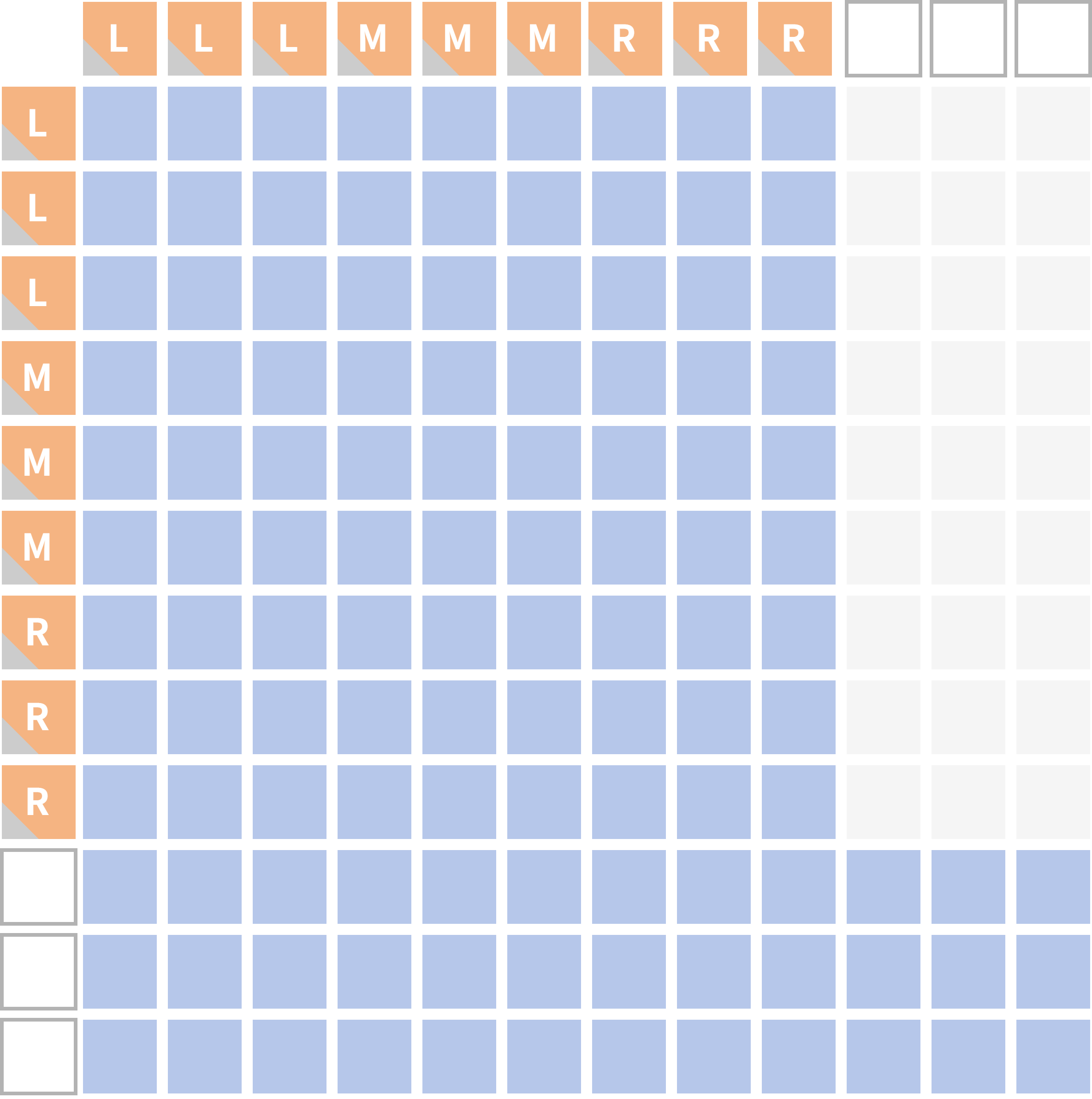}
   \caption{\textbf{Block-Wise Causal Self-Attention.} Multi-view observations $\mathcal{I} = \{M, L, R\}$ (Main, Left, Right). Compression ratio reaches 1/12. \includegraphics[height=0.9em]{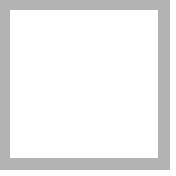} represents the Aggregation Token.}
   \label{fig:supp_1}
\end{figure}

\paragraph{CO-Fuser.} 
% Unlike VGGT and SigLIP, which fuse their outputs at the token level, we exploit the structural similarity between these encoders. Both are geometrically enhanced, non-textual semantic contrastive models, and VGGT is pretrained on large-scale 3D knowledge via DINOv2~\cite{dinov2}. This similarity enables dense, block-level fusion of the two encoders.
Unlike VGGT and SigLIP, which fuse only their final token outputs, we leverage the structural similarity between VGGT and DINOv2~\cite{dinov2} to enable a deeper form of fusion. Both encoders are geometry-enhanced, non-textual semantic contrastive models, and VGGT’s large-scale 3D pretraining is itself grounded in DINOv2. This architectural alignment makes it possible to perform dense, block-level fusion across the two encoder streams.
\begin{itemize}[leftmargin=*]
\item \textbf{Step 1 (Group-Fusion).} For the $l$-th block, we compute
\[
\mathbf{z}_{l}^{\text{geo-3D}} = (1-\alpha_l) \odot \mathbf{z}_{l}^{\text{geo}} + \alpha_l \odot \mathbf{z}_l^{\text{3D}},
\] 
where the frozen VGGT 3D features decay with layer depth following a cosine schedule, while the learned fusion features gradually gain weight through block-wise integration. In Eq. (15), $\alpha_0 = \psi = 0.2$, $\alpha_{\mathcal{L}'} = \psi \cdot \delta = 0.01$, and $\mathcal{L}'=24$.
\item \textbf{Step 2 (Aggregation Token Concatenation).} To mitigate visual ambiguities from single viewpoints by incorporating geometric relations across views, we initialize a set of aggregation tokens $\mathbf{z}_0^{\text{agg-3D}}$ and concatenate them to $\mathbf{z}_{0}^{\text{geo-3D}}$. These tokens are learned progressively through the blocks, with their number fixed at 64 (compressed to 1/8 for dual-view single-arm tasks and 1/12 for triple-view dual-arm tasks).
\item \textbf{Step 3 (IG-Aggregation).} We design a block-wise causal self-attention mechanism, illustrated in Fig.~\ref{fig:supp_1}, for IG-Aggregation. Causal attention is applied between $\mathbf{z}_{l}^{\text{geo-3D}}$ and $\mathbf{z}_l^{\text{agg-3D}}$, while bidirectional attention is used within each set. This ensures effective flow and aggregation of information from different views into $\mathbf{z}_l^{\text{agg-3D}}$. Finally, only the 64 aggregation tokens from the last block, $\mathbf{z}_{\mathcal{L}'}^{\text{agg-3D}}$, are retained.
\end{itemize}

\paragraph{CS-Thinker.} Based on $\mathbf{z}_{\{M, L, R\}}^{\text{obj-3D}}$ from CV-Aligner and $\mathbf{z}_{\mathcal{L}'}^{\text{agg-3D}}$ from CO-Fuser, CS-Thinker learns implicit semantic and geometric knowledge for reasoning via Spatiotemporal Consistency Attention, a Training-Only dynamic decoder, and depth decoders. Specifically, 
\begin{itemize}
\item \textbf{1)} three sets of initialized dynamic tokens ($3\times4=12$ tokens) are used to decode single-view dynamic object representations;
\item \textbf{2)} one set of initialized depth tokens ($1\times4=4$ tokens) is used to decode global depth across three views.
\end{itemize}
Accordingly, CS-Thinker employs 1 dynamic decoder and 3 independent depth decoders. Each decoder consists of 8 Transformer blocks with a hidden size of 1024, 16 attention heads per block, and a feed-forward ratio of 4.

\paragraph{$\alpha_l$ Settings.}
We analyze the design of the layer-wise adaptive weight $\alpha_l$ in the Group-Fusion module (Eq. 15), showing that the cosine decay mechanism is intentional. The derivative $\frac{d\alpha_l}{dl}$ approaches 0 as $l \to 0$ (shallow layers) and $l \to \mathcal{L}'$ (deep layers).
\begin{itemize}[leftmargin=*]
\item \textbf{1)} At the initial stages of training, $\alpha_l$ remains close to a high value, $\psi$, with a near-zero slope, allowing the model to focus on absorbing prior geometric information and aligning it under strong constraints.
\item \textbf{2)} In the later stages, $\alpha_l$ smoothly decreases toward its minimum value, $\psi \cdot \delta$, with the slope again approaching zero. This transition enables the model to steadily shift attention towards learned geometric features in the final layers, effectively exiting prior geometric constraints.
\item \textbf{3)} The rate of change of cosine decay is concentrated in the middle layers ($l \approx \mathcal{L}'/2$), reflecting a common trend in feature learning: models tend to focus on integrating both low-level and high-level information. The cosine decay ensures that the intensity of geometric priors primarily shifts in this region, aiding in the acceleration and completion of complex feature abstraction.
\end{itemize}
In contrast, with linear decay, the rate of change of $\alpha_l$ with respect to $l$ is constant. This results in a uniform removal of geometric priors across the entire network depth, leading to a sharp, discontinuous transition during optimization. Our $\alpha_l$ settings, with a variable rate, offer a performance advantage, as shown in Tab.~\ref{tab:supp_alpha}.

\begin{table}[t]
\caption{Comparison of the Performance between the $\alpha_l$ Settings in Eq. 15 and Linear Decay.
}
    \label{tab:supp_alpha}
    \centering
    % \small
    \footnotesize
    % \scriptsize
    % \tiny
    % \setlength{\tabcolsep}{3pt}
    \begin{tabular}{c|cc}
    \toprule
   \textbf{ $\frac{d\alpha_l}{dl}$}  & \textbf{LIBERO SR $\uparrow$} & \textbf{Real-World SR $\uparrow$}  \\
    \midrule
    \rowcolor[HTML]{DAE3F5}
      $-\frac{\psi \cdot (1-\delta)}{2} \cdot \frac{\pi}{\mathcal{L}'} \cdot \sin\left(\frac{l\pi}{\mathcal{L}'}\right)$ &     \textbf{98.1} & \textbf{78.3}  \\
      1.0 &  94.4~\textcolor{gray}{\scriptsize (-3.7)} & 73.3~\textcolor{gray}{\scriptsize (-5.0)}  \\
      0.1 & 95.9~\textcolor{gray}{\scriptsize (-2.2)} & 75.0~\textcolor{gray}{\scriptsize (-3.3)}  \\
    \bottomrule
    \end{tabular}
\end{table}

\subsection{Training Details}
\paragraph{Single-Arm Task Training.} We adopt OpenVLA~\cite{openvla} as the backbone, with an action chunk size of $K=8$. Fine-tuning is performed using low-rank adaptation (LoRA~\cite{hu2022lora}) with rank 32 and $\alpha=64$. The model is trained for 80K steps with a batch size of 64 and an initial learning rate of $5 \times 10^{-4}$. Checkpoints are evaluated every 10K steps, and the best-performing checkpoint is reported.

\paragraph{Dual-Arm Task Training.} For dual-arm tasks, the action chunk size is set to $K=25$, and OpenVLA is fine-tuned using LoRA with rank 32 and $\alpha=64$. The model is trained for 80K steps with a batch size of 32. The initial learning rate is $5 \times 10^{-4}$ and decayed to $5 \times 10^{-5}$ after 50K steps. Checkpoints are evaluated every 10K steps from step 80K onward, and the best-performing checkpoint is reported.

\section{Experimental Details}
\label{sec:experimental_details}
\subsection{Simulation Benchmark}
\paragraph{LIBERO.} LIBERO (LIfelong learning BEnchmark for RObot manipulation)~\cite{libero} is a simulation-based evaluation platform designed to investigate lifelong learning and multi-task transfer in robotic manipulation. It consists of four structured task suites, with each suite containing 10 tasks, each evaluated over 50 trials (\textbf{10 tasks $\times$ 50 trials}).

\paragraph{RoboTwin 2.0.} We randomly selected 7 different types of dual-arm tasks based on ALOHA~\cite{aloha} in RoboTwin 2.0~\cite{chen2025robotwin} to test the model's robust dual-arm manipulation ability, conducting 100 trials for each task (\textbf{7 tasks $\times$ 100 trials}):
\begin{itemize}[leftmargin=*]
\item \textbf{Click Alarmclock:} Click the center of the top-side button on the alarm clock on the table.
\item \textbf{Turn Switch:} Use the robotic arm to flip the switch.
\item \textbf{Put Bottles Dustbin:} Use the arms to grab the bottles and place them in the dustbin to the left of the table.
\item \textbf{Open Laptop:} Use one arm to open the laptop.
\item \textbf{Press Stapler:} Use one arm to press the stapler.
\item \textbf{Place Empty Cup:} Use one arm to place the empty cup on the coaster.
\item \textbf{Blocks Ranking RGB:} Place the red, green, and blue blocks in the order of red, green, and blue from left to right, placing them in a row.
\end{itemize}

\subsection{Real-World Setup}
\label{real_world_setup}
\paragraph{Experimental Platforms.} We validate the real-world performance of ConsisVLA-4D on two advanced mobile manipulation platforms: the AgileX Cobot Magic~\cite{aloha} and the Galaxea R1 Lite~\cite{galaxea}.
\begin{itemize} 
\item The AgileX Cobot Magic platform, developed by AgileX Robotics based on Stanford's ALOHA project, integrates a differential drive AGV chassis (Tracer), a four-arm collaborative system, and RGB-D sensors.
\item The Galaxea R1 Lite platform, developed by Galaxea Dynamics, is a modular dual-arm mobile platform designed for data collection and embodied intelligence development. It features a 23-DOF configuration (6-DOF chassis, 3-DOF torso, 7-DOF single-arm with gripper), an omnidirectional chassis with three steering wheels, and a suite of perception modules, including binocular cameras, a monocular depth camera, and LiDAR.
\end{itemize}

\paragraph{Task Description \& Evaluation Protocol.}
In addition to the tasks reported in the main text (Task 1 through Task 4: Microwave Operation, Banana Peeling, Drawer Arrangement, and T-shirt Folding). Tasks 1 through 4 collected \textbf{60}, \textbf{60}, \textbf{60}, and \textbf{45} expert demonstrations, respectively. ConsisVLA-4D exhibited consistently strong performance across all tasks.
\begin{itemize}
    \item \textbf{Task 1}: \textit{``\textbf{Put the bread into a bowl and heat it in the microwave.}''} 
    
   \item \textbf{Task 2}: \textit{``\textbf{Peel the banana and place it on the plate.}''} 

    \item \textbf{Task 3}: \textit{``\textbf{Open the drawer, put the toy inside, and then close the drawer.}''} 
    
    \item \textbf{Task 4}: \textit{``\textbf{Fold the T-shirt.}''} 
    
\end{itemize}

\begin{figure*}[t]
  \centering
   \includegraphics[width=1.0\linewidth]{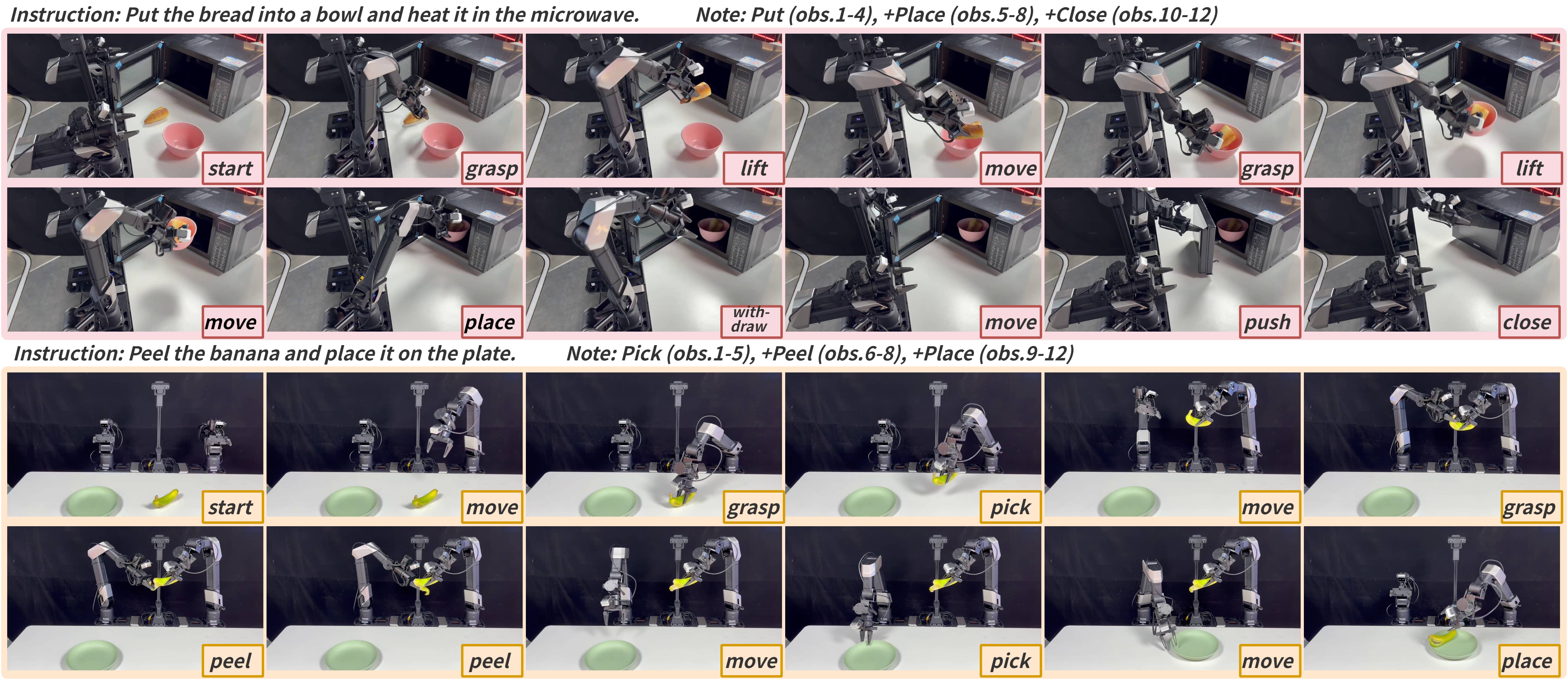}
   % \vspace{-1.8em}
   \caption{\textbf{Visualization of Task 1 and Task 2 Execution}, illustrating key execution-stage observations in full.}
   % \vspace{-12pt}
   \label{fig:supp_task12}
\end{figure*}

\section{Supplementary Visualizations}
\label{sec:supp_vis}
\subsection{Task Execution Visualizations}
As shown in Fig.~\ref{fig:supp_task12}, we present visualizations of key execution-stage observations for Tasks 1-2 in real-world settings. Tasks 1 and 2 assess the model's long-range task execution capability in complex environments, its alignment with multi-stage semantic instructions, and its precise understanding of spatial and geometric relationships.

\begin{itemize}[leftmargin=*]
\item In \textbf{Task 1} (Microwave Operation), the robot successfully executes a long sequence of actions involving object nesting and spatial constraints: it first places the bread accurately into the bowl, then smoothly inserts the bowl into the narrow interior of the microwave, and finally closes the microwave door.
\item In \textbf{Task 2} (Banana Peeling), the model demonstrates exceptional dual-arm coordination. The robot uses one arm to stabilize the object while the other arm performs the delicate peeling operation, successfully placing the peeled banana onto a plate.
\end{itemize}

\subsection{Additional CV-Aligner Visualization Results}
\begin{figure*}[t]
  \centering
   \includegraphics[width=1.0\linewidth]{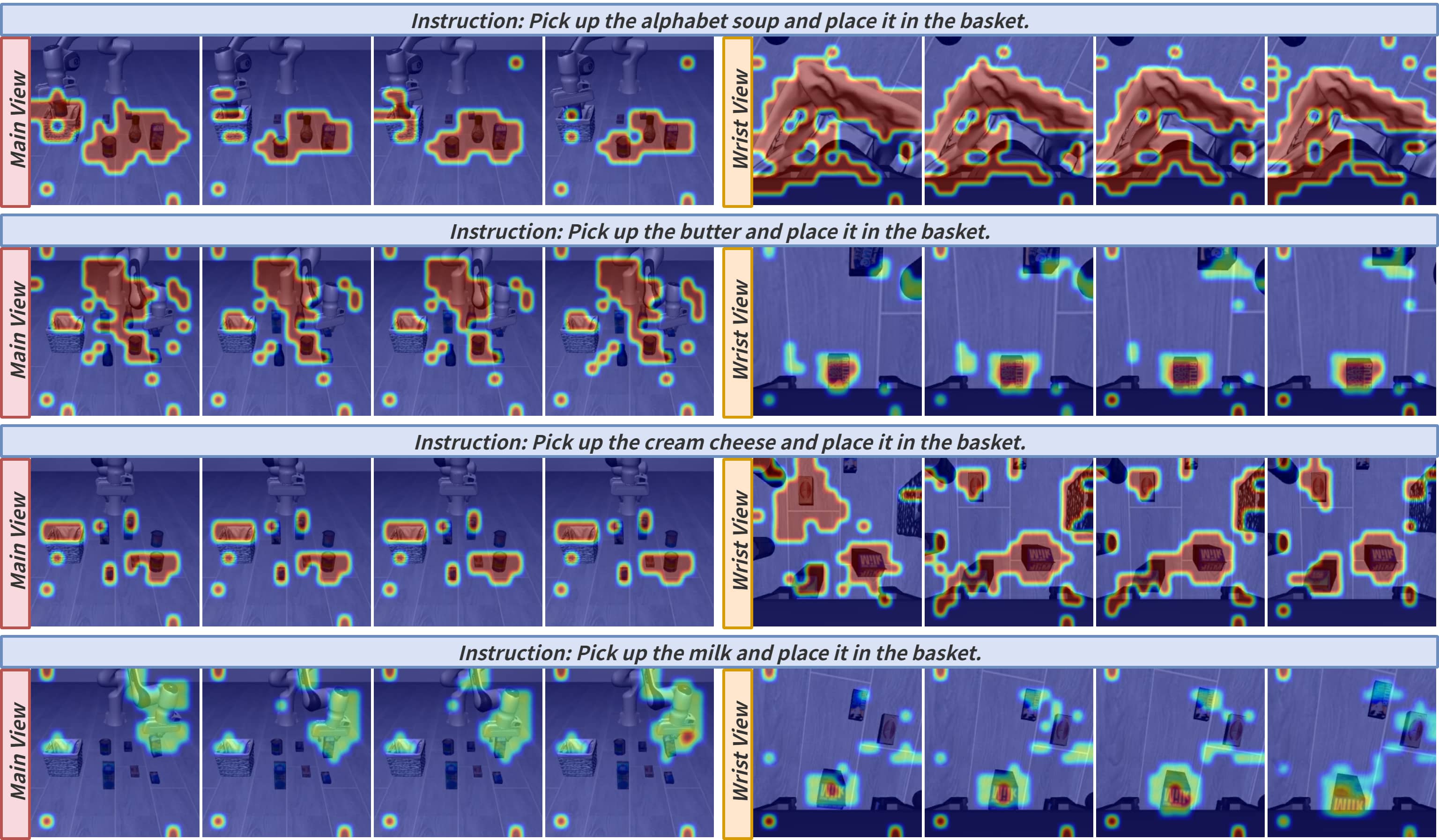}
   % \vspace{-1.8em}
   \caption{\textbf{Additional Qualitative Visualizations of CV-Aligner.} This figure illustrates the attention heatmaps generated by the CV-Aligner module in the Main View and Wrist View under different language instructions.}
   % \vspace{-12pt}
   \label{fig:supp_vis1}
\end{figure*}

As shown in Fig.~\ref{fig:supp_vis1}, we present additional qualitative visualizations of CV-Aligner. Four specific instruction cases were selected, asking the robot to pick up alphabet soup, butter, cream cheese, and milk. Each row corresponds to a particular text instruction, with attention heatmaps unfolding over time steps in both the Main View and Wrist View. It is evident that CV-Aligner effectively filters out target objects that highly match the instruction semantics from redundant background objects, validating the effectiveness of the visual redundancy removal via the Top-K selection mechanism in Equations (11-12).

\begin{figure*}[t]
  \centering
   \includegraphics[width=1.0\linewidth]{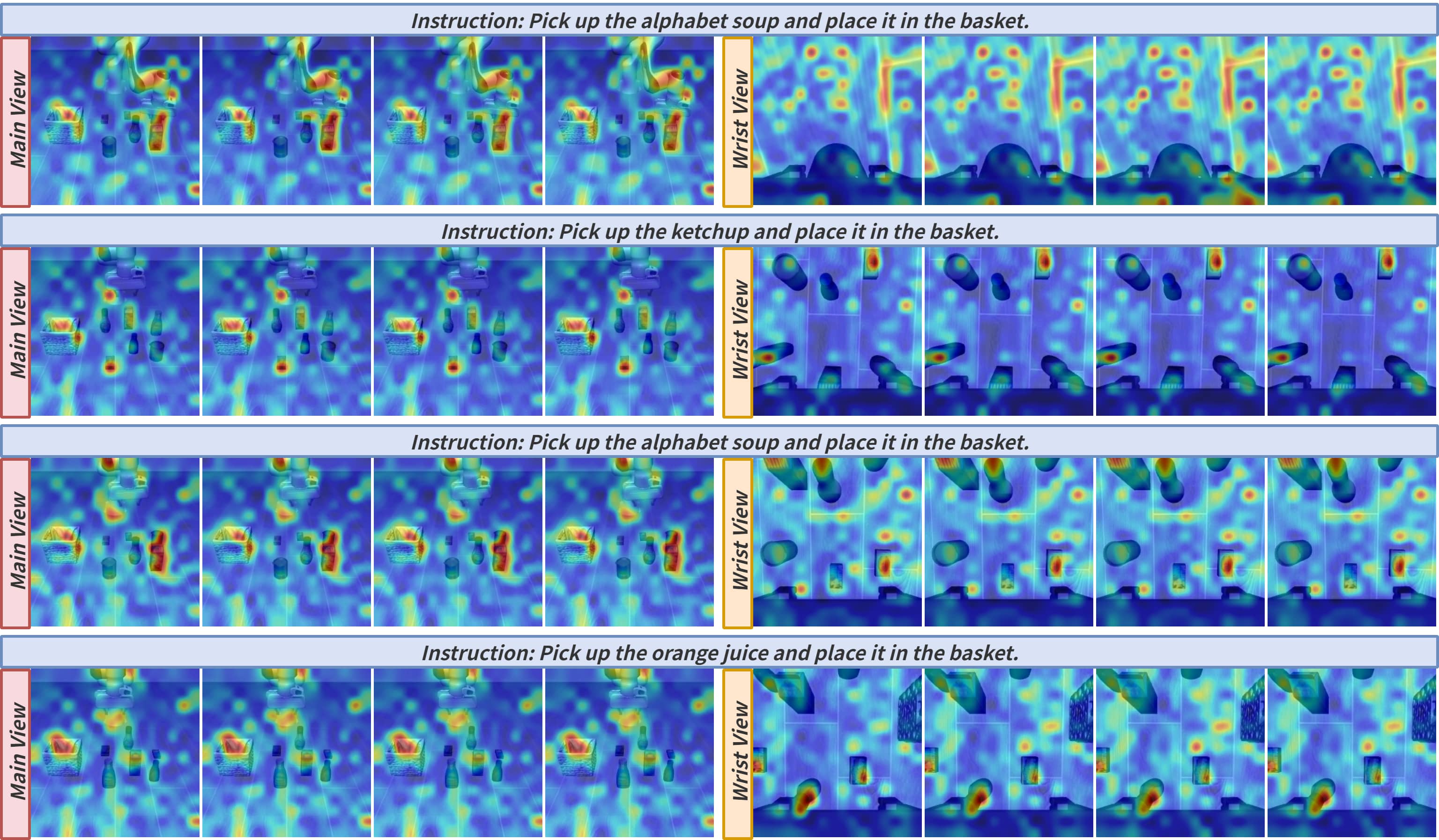}
   % \vspace{-1.8em}
   \caption{\textbf{Additional Qualitative Visualizations of CO-Fuser.} This figure illustrates the attention heatmap between the Aggregation Tokens extracted by the CO-Fuser module and the original visual patch tokens. Unlike the single-point focus of CV-Aligner, CO-Fuser presents a distributed attention pattern, complementing the focus of CV-Aligner on instruction-relevant objects.}
   % \vspace{-12pt}
   \label{fig:supp_vis2}
\end{figure*}

\subsection{Additional CO-Fuser Visualization Results}
As shown in Fig.~\ref{fig:supp_vis2}, which differs notably from Fig.~\ref{fig:supp_vis1}, the attention heatmaps of CO-Fuser illustrate how the model understands the global geometric layout of the scene and the spatial relationships between objects. Unlike focusing solely on a single object, CO-Fuser covers multiple spatial nodes relevant to the task. CO-Fuser initializes a set of Aggregation Tokens that account for only 1/12 to 1/8 of the original patch tokens, and through this set, implicitly captures the geometric relationships across all viewpoints, integrating discrete object features into coherent spatial structure information. This complements the functionality of CV-Aligner. 

Comparing the heatmaps of the Main View and Wrist View, we observe that CO-Fuser constructs a more robust spatial representation by leveraging the complementary nature of multi-view information. The Main View provides global context on the relative positions of objects (e.g., distance between an object and a basket), while the Wrist View supplements with close-range geometric details. This cross-view geometric fusion effectively addresses spatial localization uncertainty in a single viewpoint.